\documentclass[10pt,twocolumn,letterpaper]{article}
\usepackage[accsupp]{axessibility}
\usepackage{iccv}
\usepackage{times}
\usepackage{graphicx}
\usepackage{amsmath}
\usepackage{amssymb}
\usepackage{cases}
\usepackage{setspace}
\usepackage{overpic}
\usepackage{rotating}
\graphicspath{{figures/}}
\usepackage[sort,numbers]{natbib}
\usepackage{subeqnarray}
\usepackage{multirow}
\usepackage{tabularx}
\usepackage{epstopdf}
\usepackage{subfigure}
\usepackage{enumerate}
\usepackage{here}
\usepackage{bm}
\usepackage{color}
\usepackage{xcolor}
\usepackage{colortbl,booktabs}

\usepackage[pagebackref=true,breaklinks=true,letterpaper=true,colorlinks,bookmarks=false]{hyperref}

\iccvfinalcopy 

\def\iccvPaperID{} 

\begin{document}

\title{Practical Blind Image Denoising via Swin-Conv-UNet and Data Synthesis}

\author{Kai Zhang$^{1,}$\thanks{Corresponding author, email: cskaizhang@gmail.com \\This paper is accepted by Machine Intelligence Research.} \qquad Yawei Li$^1$\qquad Jingyun Liang$^{1}$\qquad Jiezhang Cao$^{1}$ \qquad Yulun Zhang$^{1}$  \\ Hao Tang$^{1}$ \qquad Deng-Ping Fan$^{1}$  \qquad   Radu Timofte$^2$  \qquad Luc Van Gool$^{1,3}$\\
$^1$CVL, ETH Z\"urich, Switzerland 
\qquad $^2$University of W\"urzburg, Germany
\qquad $^3$KU Leuven, Belgium\\
\url{https://github.com/cszn/SCUNet}
}

\maketitle

\begin{abstract}
While recent years have witnessed a dramatic upsurge of exploiting deep neural networks toward solving image denoising, existing methods mostly rely on simple noise assumptions, such as additive white Gaussian noise (AWGN), JPEG compression noise and camera sensor noise, and a general-purpose blind denoising method for real images remains unsolved. In this paper, we attempt to solve this problem from the perspective of network architecture design and training data synthesis. Specifically, for the network architecture design, we propose a swin-conv block to incorporate the local modeling ability of residual convolutional layer and non-local modeling ability of swin transformer block, and then plug it as the main building block into the widely-used image-to-image translation UNet architecture. For the training data synthesis, we design a practical noise degradation model which takes into consideration different kinds of noise (including Gaussian, Poisson, speckle, JPEG compression, and processed camera sensor noises) and resizing, and also involves a random shuffle strategy and a double degradation strategy. Extensive experiments on AGWN removal and real image denoising demonstrate that the new network architecture design achieves state-of-the-art performance and the new degradation model can help to significantly improve the practicability. We believe our work can provide useful insights into current denoising research. 
\end{abstract}

\section{Introduction}
Image denoising, which is the process of recovering a latent clean image $\mathbf{x}$ from its noisy observation $\mathbf{y}$, is perhaps the most fundamental image restoration problem.
The reason is at least three-fold. First, it can help to evaluate the effectiveness of different image priors and optimization algorithms~\cite{chatterjee2009denoising}. Second, it can be plugged into variable splitting algorithms (\eg, half-quadratic splitting~\cite{afonso2010fast} and alternating direction method of multipliers~\cite{boyd2011distributed}) to solve other problems (\eg, deblurring and super-resolution)~\cite{kamilov2017plug}. Third, it could be the very first step for other vision tasks.

The degradation model of image denoising can be mathematically formulated by
\begin{equation}\label{eq:degradation_model}
\mathbf{y} = \mathbf{x} + \mathbf{n},
\end{equation}
where $\mathbf{n}$ is the noise to be removed. Recently, deep neural networks have become the mainstream method for image denoising. To improve deep image denoising performance, researchers mainly focus on two research directions. The first one is to improve the performance under the assumption that $\mathbf{n}$ is additive white Gaussian noise (AWGN). The second one largely focuses on training data or noise modeling.
Both directions can contribute to the ultimate goal of improving the practicability for real images.

The common assumptions of $\mathbf{n}$ are AWGN, JPEG compression noise, Poisson noise, and camera sensor noise, among which AWGN is the most widely-used one due to its mathematical convenience. 
However, it is known that deep image denoising model trained by AWGN performs poorly for most of real images due to noise assumption mismatch~\cite{plotz2017benchmarking,brooks2019unprocessing}. Nevertheless, AWGN removal is fair to test the effectiveness of different network architecture designs. In recent years, various network architecture designs have been proposed. Some representative ones are DnCNN~\cite{zhang2017beyond}, N$^{3}$Net~\cite{plotz2018neural}, NLRN~\cite{liu2018non}, DRUNet~\cite{zhang2021plug}, and SwinIR~\cite{liang2021swinir}.
Indeed, network architecture designs can help to capture image prior for better image denoising performance. For example, N$^{3}$Net~\cite{plotz2018neural} and NLRN~\cite{liu2018non} are specifically designed to capture non-local image prior. Although the PSNR performance on benchmark datasets has been largely improved, \eg, SwinIR~\cite{liang2021swinir} outperforms DnCNN~\cite{zhang2017beyond} by an average PSNR of 0.57dB on Set12 dataset for noise level 25, \textit{\textbf{it is still interesting to raise the first question whether the PSNR performance can be further improved by advanced network architecture design}}.

In order to facilitate the practicability of deep denoising models, a flurry of work has been devoted to noise modeling. The motivation behind this is to make the noise assumption consistent with the degradation of real images.
Plotz and Roth~\cite{plotz2017benchmarking} establish a realistic Darmstadt Noise Dataset (DND) with consumer cameras which is composed of different pairs of real noisy and almost noiseless reference images in the RAW domain and sRGB domain. 
They further show that the model retrained with accurate degradation can significantly outperform that with AWGN on the sRGB DND dataset~\cite{plotz2018neural}.
By leveraging the physics of digital sensors and the steps of an imaging pipeline, Brooks~\etal~\cite{brooks2019unprocessing} design a camera sensor noise synthesis method and provide an effective deep raw image denoising model.
Although the above attempts have emphasized the importance of degradation models, they mainly focus on camera sensor induced noise removal. Yet, few work has been done on training a deep model for general-purpose blind image denoising. \textit{\textbf{It is interesting to raise the second question of how to improve the training data for blind denoising.}}

We attempt to answer the above two questions with novel network architecture design and novel training data synthesis. For the network architecture design, motivated by the facts that 1) different methods for image denoising have complementary image prior modeling ability and can be incorporated to boost the performance~\cite{burger2013learning}; 2) DRUNet~\cite{zhang2021plug} and SwinIR~\cite{liang2021swinir} exploit very different network architecture designs while achieving very promising denoising performance, we propose a swin-conv block to combine the local modeling ability of residual convolutional layer~\cite{he2016identity} and non-local modeling ability of swin transformer block~\cite{liu2021swin}, and then plug it as the main building block into the UNet architecture. In order to test its effectiveness, we evaluate its PNSR performance on benchmark datasets for AWGN removal. 
Since real image noise could be introduced by other types of noise, such as JPEG compression noise, processed camera sensor noise, and be further affected by resizing, it is too complex to model with a parametric probability distribution. To resolve this problem, we propose a random shuffle of different kinds of noise (including Gaussian, Poisson, speckle, JPEG compression, and processed camera sensor noises) and resizing operations (including the commonly used bilinear and bicubic interpolations) to make a rough approximation of real image noise. 

Our contributions are listed as follows:
\begin{itemize}
\item[1)] We propose a novel denoising network by plugging novel  swin-conv blocks into multiscale UNet to boost the local and non-local modeling ability.
\item[2)] We propose a hand-designed noise synthesis model, which can be used to train a general-purpose blind image denoising model.
\item[3)] 
Our blind denoising model trained with the proposed noise synthesis model can significantly improve the practicability for real images.
\item[4)]
Our work provides a strong baseline for both synthetic Gaussian denoising and practical blind image denoising. The codes will be released upon acceptance.
\end{itemize}

\section{Related work}
\label{sec:related_work}

\subsection{Deep Blind Image Denoising}
\label{ssc:blind_image_denoising}
Compared to non-blind image denoising, where the noise type and noise level are assumed to be known, blind denoising tackles the case when the noise level of certain noise type is unknown or even the noise type is unknown. During past few years, several attempts have been made to solve the problem of deep blind denoising.
Zhang~\etal~\cite{zhang2017beyond} demonstrate that a single deep model can handle Gaussian denoising with various noise levels and can even handle JPEG compression with different quality factors and single image super-resolution with different scale factors.
Chen~\etal~\cite{chen2018image} propose to adopt generative adversarial networks (GAN) to generate noise from clean images and then construct the paired training data for subsequent training.
Guo~\etal~\cite{guo2019toward} propose a convolutional blind denoising network
(CBDNet) with a noise estimation subnetwork and then propose to train the model with realistic noise model and real-world
noisy-clean image pairs.
Krull~\etal~\cite{krull2019noise2void} propose a blind-spot network which can be trained without noisy image pairs or clean target images.
Yue~\etal~\cite{yue2019variational} propose a variational inference method for blind image denoising which incorporates both noise estimation
and image denoising into a unique Bayesian framework.
While achieving promising results, the above methods are mainly evaluated on the synthetic Gaussian noise or the processed camera sensor noise such as the DND dataset~\cite{plotz2017benchmarking}. 
Since real noise is far more complex, the above methods can not be readily applied for real applications.
It is still unclear how to establish more practical noisy/clean image pairs for training a deep blind model.

\subsection{Deep Architecture for Non-Local Prior}
\label{ssc:degradation_models}
State-of-the-art model-based image denoising methods mostly exploit non-local self-similarity (NLSS) prior
which refers to the fact that a local patch often has many non-local similar patches across the image~\cite{buades2005non}. Some representative methods include BM3D~\cite{dabov2007image}, LSSC~\cite{mairal2009non} and WNNM~\cite{gu2014weighted}.
Inspired by the effectiveness of NLSS prior, some deep learning methods attempt to explicitly model the correlation among non-local patches via the network structure.
Sun and Tappen~\cite{sun2011learning} propose a gradient-based discriminative non-local range Markov Random Field (MRF) method to exploit the advantages of BM3D and non-local means. 
Inspired by non-local variational
methods, Lefkimmiatis~\cite{lefkimmiatis2017non} designs an unrolled network
that can perform non-local processing for better denoising performance. However, the above methods adopt the non-differentiable KNN matching in fixed feature spaces. To resolve this,
Plotz and Roth~\cite{plotz2018neural} propose a fully end-to-end trainable neural nearest neighbor block to leverage the principle of non-local self-similarity. 
Liu~\etal~\cite{liu2018non} propose a non-local recurrent network (NLRN) to incorporate non-local operations into a recurrent neural network.
Chen~\etal~\cite{chen2021pre} propose image processing transformer (IPT) to exploit the non-local modeling of transformer. However, IPT works on fixed image patch size and tends to result in border artifacts.
Liang~\etal~\cite{liang2021swinir} address this issue by adopting the swin transformer as the main building block. It has been shown that transformer-based methods favors more on images with repetitive structures, which verifies the effectiveness of the transformer for non-local modeling ability.

\section{Method}
\label{method}

Since we focus on learning a deep blind model with paired training data, it is necessary to revisit the Maximum A Posteriori (MAP) inference to have a better understanding.
Generally, for the problem of blind image denoising, the estimated clean image $\hat{\mathbf{x}}$ can be obtained by solving the following MAP problem with a certain optimization algorithm,
\begin{equation}\label{eq:MAP1}
\hat{\mathbf{x}} =  \mathop{\arg\min}\displaystyle_{\mathbf{x}}D(\mathbf{x}, \mathbf{y})  + \lambda P(\mathbf{x}),
\end{equation}
where $D(\mathbf{x}, \mathbf{y})$ is the data fidelity term, $P(\mathbf{x})$ is the prior term and $\lambda$ is the trade-off parameter.

So far, one can see that the key of solving blind denoising lies in modeling the degradation process of noisy image as well as designing the image prior of clean image.

By treating the deep model as a compact unrolled inference of Eq.~\eqref{eq:MAP1}, 
the deep blind denoising generally aims to solve the following bi-level optimization problem~\cite{schmidt2014shrinkage,chen2015trainable}
\begin{subequations}\label{eq_bilevel}
\begin{numcases}{}
\min_W \sum\limits_{i}\mathcal{L}(\mathbf{\hat{x}}_i(\mathbf{y}_i, W), \mathbf{x}_i) \label{eq_bilevel_1}\\
s.t. \quad \hat{\mathbf{x}}_i =  \mathop{\arg\min}\displaystyle_{\mathbf{x}}D(\mathbf{x}, \mathbf{y}_i)  + \lambda P(\mathbf{x}), \label{eq_bilevel_2}
\end{numcases}
\end{subequations}
where $W$ denotes the network parameters to be learned, $\{\mathbf{y}_i, \mathbf{x}_{i}\}$ represents the training noisy-clean image pairs, $\mathcal{L}(\cdot)$ is the loss function. In this sense, the deep blind denoising model should capture the knowledge of degradation process and image prior.

On the other hand, the modeling ability of a deep model generally depends on network architecture, model size (or the number of parameters), and training data.
It is clear that the degradation process is implicitly defined by the noisy images from the training data, which indicates the noisy images of the training data is responsible for deep blind denoising model to capture the knowledge of degradation process.
In order to improve the image prior modeling ability of deep blind denoising model, one should focus on improving the following three factors, including network architecture, model size and clean images of the training data.
While the later two factors are easy to improve, how to improve the network architecture remains further study. 

From the above discussions and analyses, we can conclude that the network architecture and the training data are two important factors to improve the performance of deep blind denoising model. In the following, we will separately detail our attempts to improve these two factors.


\subsection{Swin-Conv-UNet}
\label{sec:hscu}

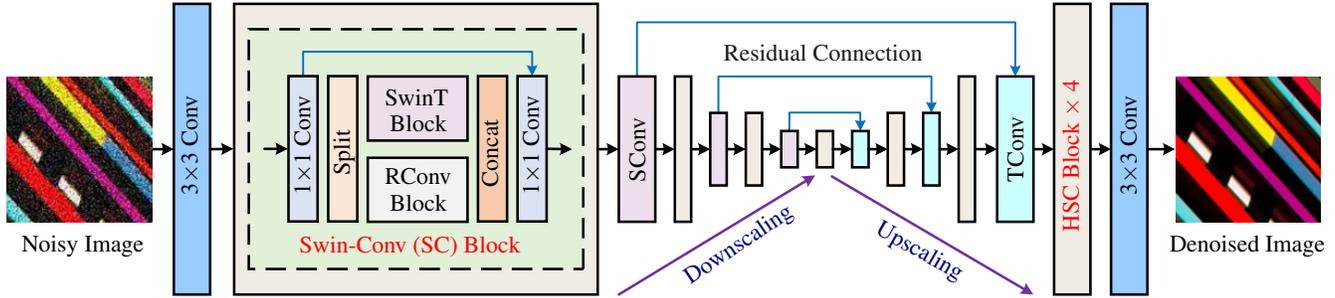
\begin{figure*}[!tbp]
\centering
\begin{overpic}[width=1\textwidth]{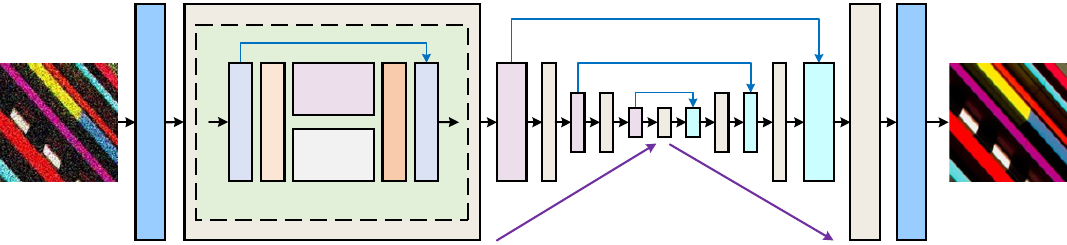}
\begin{turn}{90}
\put(8,-14.5){\color{black}{\small \textcolor[rgb]{0.00,0.00,0.00}{3$\times$3 Conv}}}
\put(8,-23.1){\color{black}{\small \textcolor[rgb]{0.00,0.00,0.00}{1$\times$1 Conv}}}
\put(9.5,-26.1){\color{black}{\small \textcolor[rgb]{0.00,0.00,0.00}{Split}}}

\put(8.6,-37.4){\color{black}{\small Concat}}
\put(8,-40.5){\color{black}{\small 1$\times$1 Conv}}

\put(9.0,-48.6){\color{black}{\small SConv}}
\put(9.0,-77.4){\color{black}{\small TConv}}
\put(5.2,-81.6){\color{black}{\small \textcolor[rgb]{1.00,0.00,0.00}{HSC Block $\times$ 4}}}
\put(8,-86){\color{black}{\small 3$\times$3 Conv}}
\end{turn}
\put(-57.4,14.8){\color{black}{\small SwinT}}
\put(-57,12.8){\color{black}{\small Block}}
\put(-57.4,8.8){\color{black}{\small RConv}}
\put(-57,6.8){\color{black}{\small Block}}
\put(-63.9,3.6){\small \textcolor[rgb]{1.00,0.00,0.00}{Swin-Conv (SC) Block}}
\put(-31.6,18.2){\small \textcolor[rgb]{0.10,0.10,0.10}{Residual Connection}}
\begin{turn}{34}
\put(-42,29.7){\color{black}{\small \textcolor[rgb]{0.00,0.00,0.50}{Downscaling}}}
\end{turn}
\begin{turn}{-30}
\put(-44,-19.5){\color{black}{\small \textcolor[rgb]{0.00,0.00,0.50}{Upscaling}}}
\end{turn}
\put(-112.8,3.7){\color{black}{\small \textcolor[rgb]{0.00,0.00,0.00}{Noisy Image}}}
\put(-25.4,3.7){\color{black}{\small \textcolor[rgb]{0.00,0.00,0.00}{Denoised Image}}}
\end{overpic}
\vspace{0.02cm}
\caption{The architecture of the proposed  Swin-Conv-UNet (SCUNet) denoising network. SCUNet exploits the swin-conv (SC) block as the main building block of a UNet backbone. In each SC block, the input is first passed through a 1$\times$1 convolution, and subsequently is split evenly into two feature map groups, each of which is then fed into a swin transformer (SwinT) block and residual 3$\times$3 convolutional (RConv) block, respectively; after that,
the outputs of SwinT block and RConv block are concatenated and then passed through a 1$\times$1 convolution to produce the residual of the input.
``SConv''  and ``TConv'' denote 2$\times$2 strided convolution with stride 2 and 2$\times$2 transposed convolution with stride 2, respectively.}
\label{network}
\end{figure*}

Fig.~\ref{network} illustrates the network architecture of our proposed  Swin-Conv-UNet (SCUNet). The main idea of SCUNet is to integrate the complementary network architecture designs of DRUNet and SwinIR. To be specific, SCUNet plugs novel  swin-conv (SC) blocks into a UNet~\cite{ronneberger2015u} backbone. Following DRUNet~\cite{zhang2021plug}, the UNet backbone of SCUNet has four scales, each of which has a residual connection between 2$\times$2 strided convolution (SConv) based downscaling and 2$\times$2 transposed convolution (TConv) based upscaling. The number of channels in each layer from the first scale to the fourth scale are 64, 128, 256 and 512, respectively. The main difference between SCUNet and DRUNet is that SCUNet adopts four SC blocks rather than four residual convolution blocks in each scale of the downscaling and upscaling.

As shown in the dashed line of Fig.~\ref{network}, an SC block fuses swin transformer (SwinT) block~\cite{liu2021swin} and residual convolutional (RConv) block~\cite{he2016identity,lim2017enhanced} via two 1$\times$1 convolutions, split and concatenation operations, and a residual connection. To be specific, for an input feature tensor $X$, it is first passed through a 1$\times$1 convolution. Subsequently, it is split evenly into two features map groups $X_1$ and $X_2$. We formulate such a process as
\begin{equation}\label{eq:splitconv}
X_1, \; X_2 = \texttt{Split}(\texttt{Conv1}\!\times\!\texttt{1}(X)).
\end{equation}
Then, $X_1$ and $X_2$ are separately fed into a SwinT block and a RConv block, giving rise to
\begin{equation}\label{eq:swinconv}
Y_1, \; Y_2 = \texttt{SwinT}(X_1), \; \texttt{RConv}(X_2).
\end{equation}
Finally, $Y_1$ and $Y_2$ are concatenated as the input of a 1$\times$1 convolution which has a residual connection with the input $X$. As such, the final output of SC block is given by
\begin{equation}\label{eq:convconcat}
Z = \texttt{Conv1}\!\times\!\texttt{1}(\texttt{Concat}(Y_1, \; Y_2)) + X.
\end{equation}

It is worth pointing out that our proposed SCUNet enjoys several merits due to some novel module designs. First, the SC block fuses the local modeling ability of RConv block and non-local modeling ability of SwinT block. Second, the local and non-local modeling ability of SCUNet is further enhanced via the multiscale UNet. Third, the 1$\times$1 convolution can effectively and efficiently facilitate information fusion between SwinT block and RConv block. Fourth, the split and concatenation operations can act as the group convolution with two groups to reduce the computation complexity and the number of parameters. 
We note that SCUNet essentially functions as a hybrid convolutional neural networks (CNNs)-Transformer network and there also exist several other works that integrate CNNs and Transformer for effective network architecture design. Li~\etal~\cite{li2021bossnas} propose a fabric-like hybrid CNN-transformer search space, in which each layer can flexibly choose CNN building blocks and transformer building blocks. Yuan~\etal~\cite{yuan2021incorporating} introduce a novel Convolution-enhanced Image Transformer (CeiT) that combines the advantages of CNNs in extracting low-level features and strengthening locality with the benefits of Transformers in establishing long-range dependencies.
Guo~\etal~\cite{guo2022cmt} propose a new Transformer-based hybrid network that leverages the strengths of Transformers to capture long-range dependencies and the capabilities of CNNs to extract local information.

It is also worth pointing out the difference between our proposed SCUNet and two recently works including Uformer~\cite{wang2021uformer} and Swin-Unet~\cite{cao2021swin}. First, the motivation is different. Our SCUNet is motivated by the fact that state-of-the-art denoising methods DRUNet~\cite{zhang2021plug} and SwinIR~\cite{liang2021swinir} exploit very different network architecture designs, and thus SCUNet
aims to integrate the complementary network architecture designs of DRUNet and SwinIR. By contrast, Uformer and Swin-Unet aim to combine the transformer variants and UNet. 
Second, the main building blocks are different. Our SCUNet adopts a novel swin-conv block which integrates the local modeling ability of residual convolutional layer~\cite{he2016identity} and non-local modeling ability of swin transformer block~\cite{liu2021swin} via 1$\times$1 convolution, split and concatenation operations. By contrast, Uformer adopts a new transformer block by combining depth-wise convolution layers~\cite{li2021localvit}, while Swin-Unet utilizes the swin transformer block as the main building block.

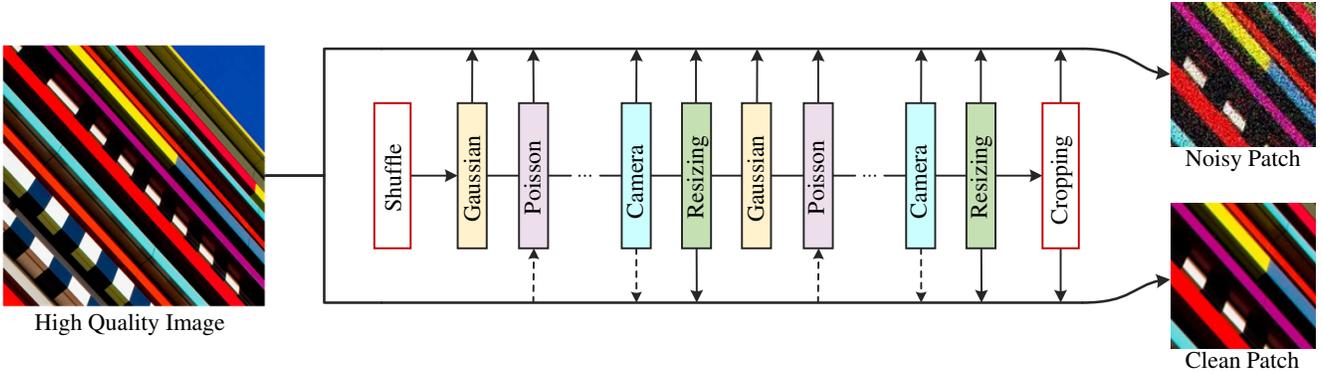
\begin{figure*}[!htbp]
\centering
\begin{overpic}[width=1\textwidth]{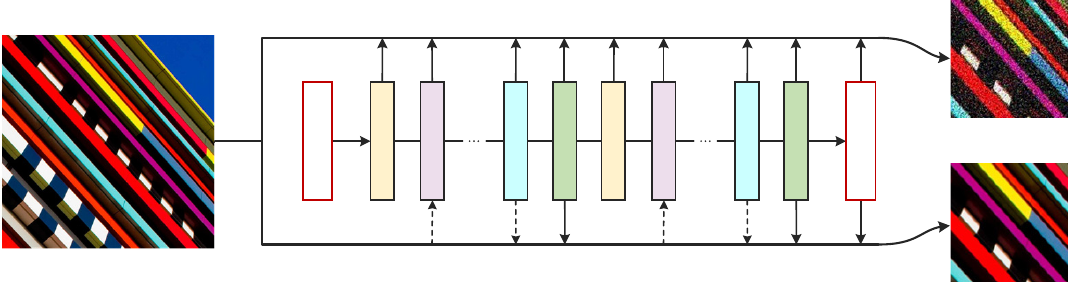}
\begin{turn}{90}
\put(10.4,-30.4){\color{black}{\small \textcolor[rgb]{0.00,0.00,0.00}{Shuffle}}}

\put(9.6,-36.4){\color{black}{\small Gaussian}}
\put(10.2,-41){\color{black}{\small Poisson}}
\put(9.9,-48.8){\color{black}{\small Camera}}
\put(9.9,-53.4){\color{black}{\small Resizing}}

\put(9.6,-58){\color{black}{\small Gaussian}}
\put(10.2,-62.6){\color{black}{\small Poisson}}
\put(9.9,-70.4){\color{black}{\small Camera}}
\put(9.9,-75.0){\color{black}{\small Resizing}}

\put(9.4,-81.1){\color{black}{\small Cropping}}
\end{turn}
\put(-79,1.4){\color{black}{\small High Quality Image}}
\put(8.6,13.9){\color{black}{\small Noisy Patch}}
\put(8.6,-1.5){\color{black}{\small Clean Patch}}

\end{overpic}

\vspace{0.3cm}
\caption{Schematic illustration of the proposed paired training patches synthesis pipeline. For a high quality image, a randomly shuffled degradation sequence is performed to produce a noisy image. Meanwhile, the resizing and reverse-forward tone mapping are performed to produce a corresponding clean image. Paired noisy/clean training patches are then cropped for training deep blind denoising model. Note that, since Poisson noise is signal-dependent, the dashed arrow for ``Poisson'' means the clean image is used to generate the Poisson noise. To tackle the color shift issue, the dashed arrow for ``Camera Sensor'' means the reverse-forward tone mapping is performed on the clean image.}
\label{degradation}
\end{figure*}

\subsection{Training Data Synthesis}
\label{sec:datasynthesis}

Instead of establishing a large variety of real noisy/clean image pairs, which is laborious and challenging, we attempt to synthesize realistic noisy/clean image pairs. The main idea is to add different kinds of noise and also include the resizing, as well as incorporating a double degradation strategy and a random shuffle strategy which we will detail next. 

\begin{spacing}{1.5}
\end{spacing}
\noindent
\textbf{Gaussian noise.} Additive white Gaussian noise (AWGN) is the most widely-used assumption for denoising. While it can perfectly model read noise of an image sensor, it usually does not match the real noise and would deteriorate the practicability of trained deep denoising models. 
Nevertheless, it has been shown that deep denoising model (\eg, FFDNet~\cite{zhang2018ffdnet}) trained with AWGN can remove non-Gaussian noise by setting a larger Gaussian noise level, with the sacrifice of smoothing the textures and edges. 
Instead of using the simplified AWGN, we adopt the 3D generalized zero-mean Gaussian noise model~\cite{nam2016holistic} with $3\times3$ covariance matrix to model the noise correlation between R, G, and B channels. One of the underlying reasons is that the color image demosaicing step in camera ISP pipeline can correlate the noise across channels. Depending on the cross-channel dependencies, such a generalized Gaussian model has two extreme cases, including the widely-used additive white Gaussian color noise and grayscale Gaussian noise. We uniformly sample their noise levels from $\{2/255,3/255, \cdots, 50/255\}$. Since in-camera denoising algorithms generally remove the color noise for better perceptual quality, grayscale Gaussian noise would be a good choice to model the remaining noise. For this reason, we sample the two extreme cases and general case with probabilities 0.4, 0.4 and 0.2, respectively.

\begin{spacing}{1.5}
\end{spacing}
\noindent
\textbf{Poisson noise.} 
Poisson noise generally refers to the photon shot noise which originates from the discrete nature of electric charge. It occurs severely in low-light conditions, such as night time photography, medical imaging, optical microscopy imaging and astronomy imaging~\cite{hasinoff2014photon}. Different from Gaussian noise which is signal-independent, Poisson noise is signal-dependent.
Traditional model-based methods mostly apply the variable-stabilizing transformation (VST) to transfer the noise into approximate signal-independent one, and then tackle the problem with Gaussian denoising methods. However, such methods need to know the noise type beforehand, which is generally impossible for real images. Hence, removing the Poisson noise directly via the deep model would be a good choice. To sample different noise levels for an image, we first multiply the clean image by $10^\alpha$, where $\alpha$ is uniformly chosen from $[2, 4]$, and then divide back by $10^\alpha$ after adding the signal-dependent Poisson noise. Our Poisson noise can be mathematically modeled as
\begin{equation}\label{eq:poisson}
\mathbf{n} \, \sim \, \mathcal{P}(10^\alpha \cdot \mathbf{x}) / 10^\alpha - \mathbf{x}; \quad \alpha \, \sim \, \mathcal{U}(2, 4).
\end{equation}
Following the Gaussian noise, we also consider grayscale Poisson noise by converting the clean color image into grayscale image.
After that, we add the same grayscale noise to each channel of the given image.

\begin{spacing}{1.5}
\end{spacing}
\noindent
\textbf{Speckle noise.} 
Speckle noise is multiplicative noise which usually appears in coherent imaging systems, such as synthetic aperture
radar (SAR) imaging and medical ultrasonic imaging~\cite{tur1982speckle,racine1999speckle}. It can be modeled by the multiplication between latent clean image and Gaussian noise. We thus simply modify the above Gaussian noise synthesis strategy by multiplying the clean image to generate speckle noise.

\begin{figure*}[!htbp]\footnotesize
\begin{tabular}{c@{\extracolsep{0em}}@{\extracolsep{0.04em}}c@{\extracolsep{0.04em}}c@{\extracolsep{0.04em}}c@{\extracolsep{0.04em}}@{\extracolsep{0.04em}}c@{\extracolsep{0.04em}}c}
         \includegraphics[width=0.163\textwidth]{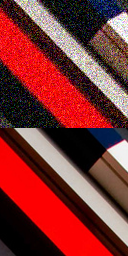}~
		&\includegraphics[width=0.163\textwidth]{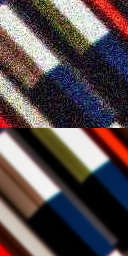}~
		&\includegraphics[width=0.163\textwidth]{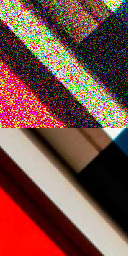}~
		&\includegraphics[width=0.163\textwidth]{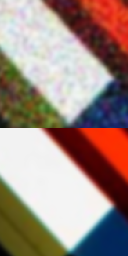}~
		&\includegraphics[width=0.163\textwidth]{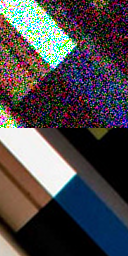}~
		&\includegraphics[width=0.163\textwidth]{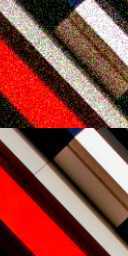}\\

	\end{tabular}
 \vspace{0.2cm}
	\caption{Synthesized noisy/clean patch pairs via our proposed training data synthesis pipeline. The size of the high quality image patch is 544$\times$544. The size of the noisy/clean patches is 128$\times$128.}
	\label{fig:syn}
\end{figure*}

\begin{spacing}{1.5}
\end{spacing}
\noindent
\textbf{JPEG compression noise.} 
Image compression can help to reduce the storage and bandwidth requirements for digital images. Among various image compression standards and formats, JPEG has been the most widely-used one since it is simple and allows for fast encoding and decoding. However, it will reduce the image quality by introducing severe 8$\times$8 blocking artifacts with the increase of compression degree. Such a trade-off is controlled by the quality factor which ranges from 0 to 100. Due to its pervasiveness in Internet and social media usage, we add this kind of noise by uniformly sampling the quality factor from $[20, 95]$.

\begin{spacing}{1.4}
\end{spacing}
\noindent
\textbf{Processed camera sensor noise.} 
The noise in output RGB image of modern digital cameras is mainly caused by passing the read and shot noise in raw sensor data through an image signal processing (ISP) pipeline. 
Hence, the distribution of the processed camera sensor noise varies with the read and shot noise model and ISP model.
Inspired by~\cite{brooks2019unprocessing}, we synthesize this kind of noise by generating raw image from clean image via the reverse ISP pipeline, and then processing the noisy raw image via the forward ISP pipeline after adding read and shot noise to raw image.
For the read and shot noise model, we exactly borrow the one proposed in~\cite{brooks2019unprocessing}.
For the ISP model, we adopt the one proposed in~\cite{zhang2021designing} which consists of demosaicing, exposure compensation, white balance, camera to XYZ (D50) color space conversion, XYZ (D50) to linear RGB color space conversion, tone mapping and gamma correction.
It is still worth pointing out the following details about the adopted ISP model. 
First, the orders of gamma correction and tone mapping, and and the tone mapping curves are different from these in~\cite{brooks2019unprocessing}. Our ISP adopts gamma correction as the final step, whereas~\cite{brooks2019unprocessing} uses  tone mapping as the final step. 
While it has been known that the tone mapping curves for different cameras are usually different, \cite{brooks2019unprocessing} uses a fixed tone curve. By contrast, our ISP selects the best fitted tone curves from~\cite{grossberg2003space} for each camera based on the error between reconstructed output and the camera ground-truth RGB output. 
Second, the forward-reverse tone mapping may cause color shift issue with respect to the original image due to the irreversibility, we resolve this by also applying the reverse-forward tone mapping for the clean image.

\begin{spacing}{1.5}
\end{spacing}
\noindent
\textbf{Resizing.} 
Image resizing is one of the basic digital image editing tools. It can be used to fit the image into a certain space on a screen or be used to downscale the image to reduce the storage size. While resizing does not introduce noise to the clean images, it would affect the noise distribution of the noisy images.
For example, upscaling would lead AWGN to be spatially correlated while downscaling would change processed camera sensor noise to be less signal-dependent.
To model such resizing induced noise, we uniformly apply the widely-used bilinear and bicubic resizing operations. The scaling factor is uniformly chosen from $[0.5, 2]$. \textbf{\textit{Especially noteworthy is that we apply the same resizing on both noisy image and its clean counterpart since resizing will change the spatial resolution of latent clean image of the noisy image}}. Hence, it is essentially different from the super-resolution degradation proposed in~\cite{zhang2021designing,wang2021real}.

\begin{table*}[!htbp]\footnotesize
\caption{Average PSNR(dB) results of different methods for grayscale image denoising with noise levels 15, 25 and 50 on the widely-used Set12, BSD68 and Urban100 datasets. The best and second best results are highlighted in \textcolor[rgb]{1.00,0.00,0.00}{red} and \textcolor[rgb]{0.00,0.00,1.00}{blue} colors, respectively.} 
\center
\begin{tabular}{p{0.8cm}<{\centering}p{1.0cm}<{\centering}|p{0.8cm}<{\centering}p{0.7cm}<{\centering}p{0.7cm}<{\centering}p{0.8cm}<{\centering}p{0.7cm}<{\centering}p{0.7cm}<{\centering}p{0.8cm}<{\centering}p{0.8cm}<{\centering}p{0.8cm}<{\centering}p{1cm}<{\centering}p{1cm}<{\centering}}
  \toprule
 \multirow{2}{*}{Datasets} & Noise  & \multirow{2}{*}{DnCNN} & \multirow{2}{*}{FFDNet} & \multirow{2}{*}{$\text{N}^3$Net}  & \multirow{2}{*}{NLRN}& \multirow{2}{*}{RNAN}& \multirow{2}{*}{FOCNet}  & \multirow{2}{*}{DAGL} &\multirow{2}{*}{DRUNet} & \multirow{2}{*}{SwinIR}& \multirow{2}{*}{Restormer} & \multirow{2}{*}{\textbf{SCUNet}}\\

 & Level &  &   & &   & &  &  &  &    &    \\ \hline\hline
      & 15 & 32.86  & 32.75  & --    & {33.16} & --      & 33.07    & 33.28  & {33.25}     & 33.36 & \textcolor[rgb]{0.00,0.00,1.00}{33.42} & \textcolor[rgb]{1.00,0.00,0.00}{33.43}  \\
Set12 & 25 & 30.44  & 30.43  & 30.55 & {30.80} & --      & 30.73    & 30.93  & {30.94}     & 31.01 & \textcolor[rgb]{0.00,0.00,1.00}{31.08} &  \textcolor[rgb]{1.00,0.00,0.00}{31.09} \\
      & 50 & 27.18  & 27.32  & 27.43 & 27.64   & {27.70} & 27.68    & 27.81  & {27.90}     & 27.91 & \textcolor[rgb]{0.00,0.00,1.00}{28.00} & \textcolor[rgb]{1.00,0.00,0.00}{28.04}  \\\hline

       & 15 & 31.73  & 31.63  & --    & {31.88} & --    & 31.83     & 31.93 & {31.91}    & \textcolor[rgb]{0.00,0.00,1.00}{31.97}& {31.96} & \textcolor[rgb]{1.00,0.00,0.00}{31.99} \\
 BSD68 & 25 & 29.23  & 29.19  & 29.30 & {29.41} & --    & 29.38     & 29.46 &{29.48}     & 29.50& \textcolor[rgb]{0.00,0.00,1.00}{29.52} & \textcolor[rgb]{1.00,0.00,0.00}{29.55} \\
       & 50 & 26.23  & 26.29  & 26.39 & 26.47   & 26.48 & {26.50}   & 26.51 &{26.59}     & 26.58 & \textcolor[rgb]{0.00,0.00,1.00}{26.62} & \textcolor[rgb]{1.00,0.00,0.00}{26.67} \\\hline
  
        & 15 & 32.64  & 32.40  & --    & {33.45}& --    & 33.15   & \textcolor[rgb]{0.00,0.00,1.00}{33.79} & 33.44 & 33.70& \textcolor[rgb]{0.00,0.00,1.00}{33.79} & \textcolor[rgb]{1.00,0.00,0.00}{33.88} \\
Urban100& 25 & 29.95  & 29.90  & 30.19 & 30.94  & --    & 30.64   & 31.39 & 31.11 & 31.30& \textcolor[rgb]{0.00,0.00,1.00}{31.46} & \textcolor[rgb]{1.00,0.00,0.00}{31.58} \\
        & 50 & 26.23 & 26.50  & 26.26 & 27.49  & 27.65 & 27.40   & 27.97 & 27.96 & 27.98  &\textcolor[rgb]{0.00,0.00,1.00}{28.29} & \textcolor[rgb]{1.00,0.00,0.00}{28.56} \\\bottomrule
\end{tabular}
\label{table_denoising_grayscale}
\end{table*}

\begin{figure*}[!htbp]\small
\hspace{-0.17cm}
\begin{tabular}{c@{\extracolsep{0em}}c@{\extracolsep{0.04em}}c@{\extracolsep{0.04em}}c@{\extracolsep{0.04em}}c@{\extracolsep{0.04em}}@{\extracolsep{0.04em}}c@{\extracolsep{0.04em}}c}

        \includegraphics[width=0.16\textwidth]{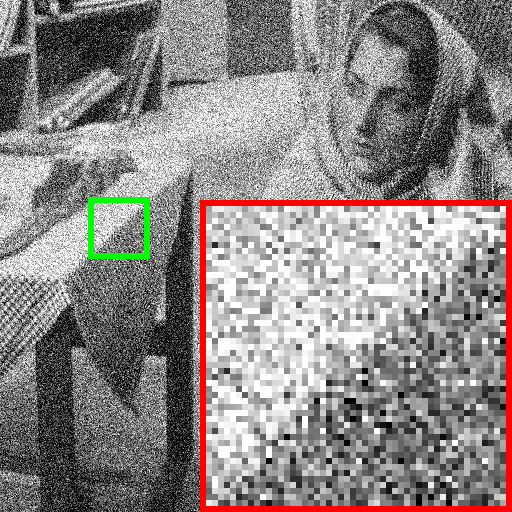}~
		&\includegraphics[width=0.16\textwidth]{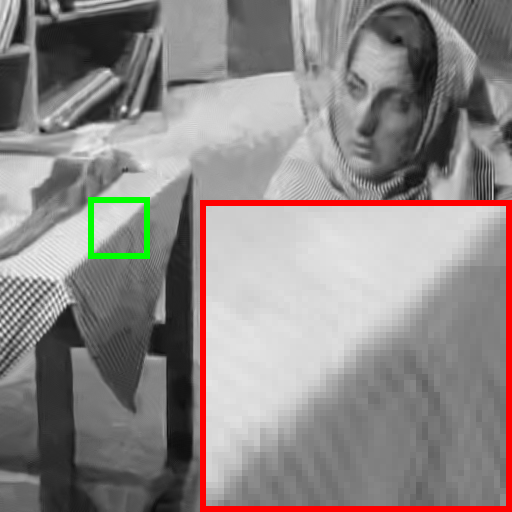}~
		&\includegraphics[width=0.16\textwidth]{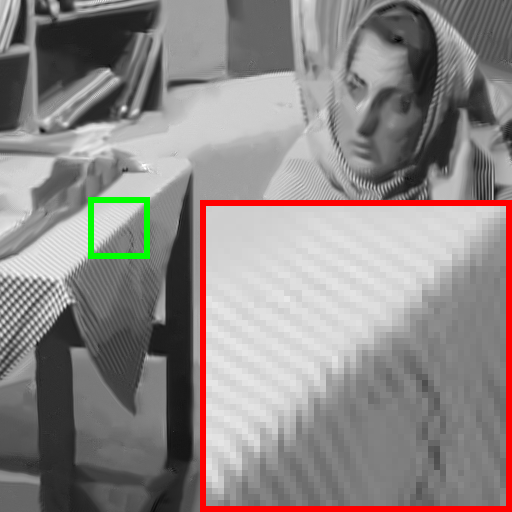}~
        &\includegraphics[width=0.16\textwidth]{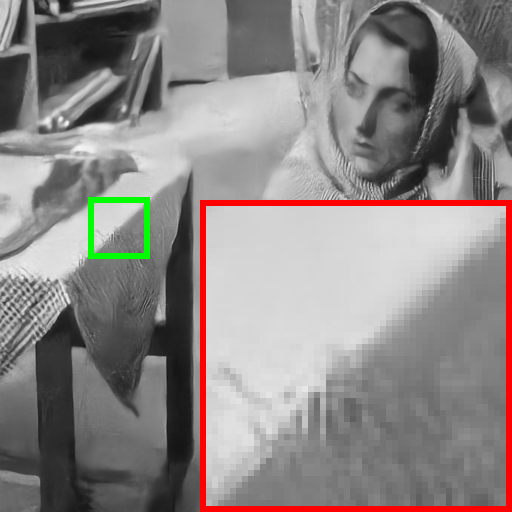}~
		&\includegraphics[width=0.16\textwidth]{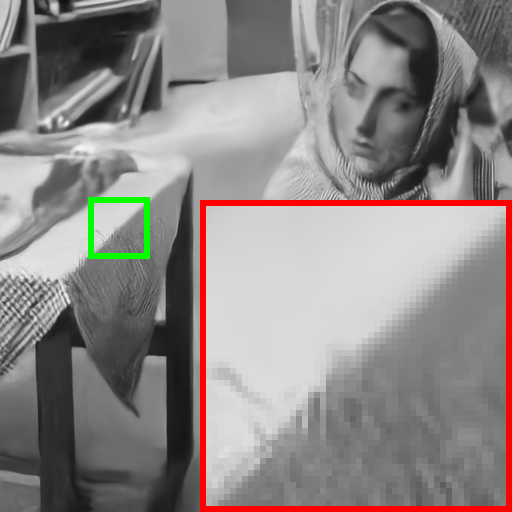}~
		&\includegraphics[width=0.16\textwidth]{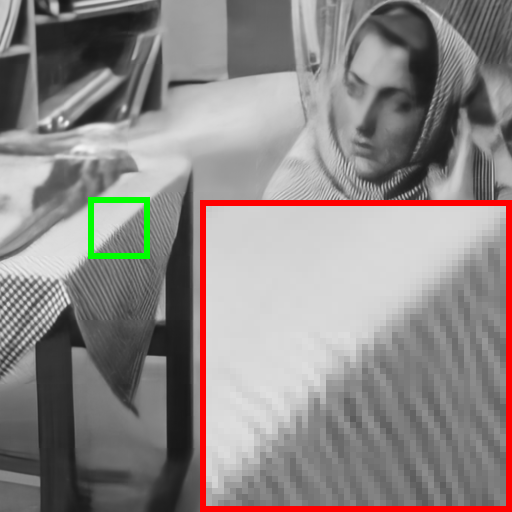}\\
 PSNR(dB)/SSIM &27.22/0.7942 & 27.79/0.8196  & 26.15/0.7682 & 26.41/0.7792 & 26.67/0.8220 \\
(a) Noisy & (b) BM3D  & (c) WNNM & (d) DnCNN & (e) FFDNet &  (f) RNAN \\

        \includegraphics[width=0.16\textwidth]{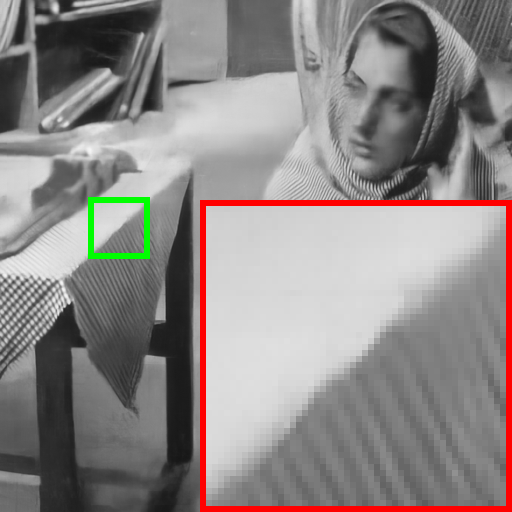}~
		&\includegraphics[width=0.16\textwidth]{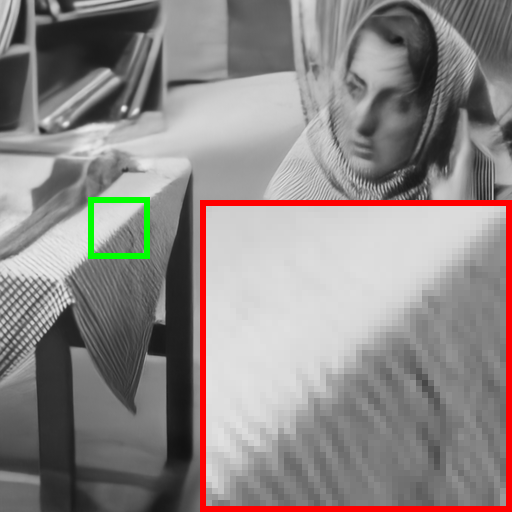}~
		&\includegraphics[width=0.16\textwidth]{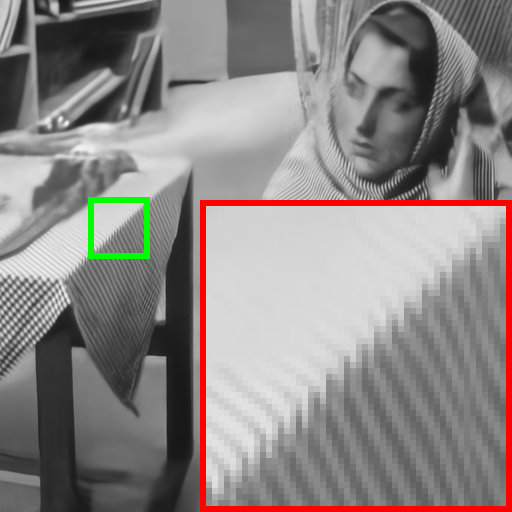}~
        &\includegraphics[width=0.16\textwidth]{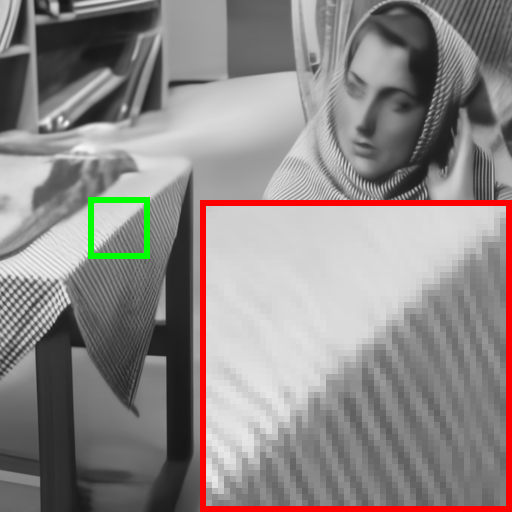}~
		&\includegraphics[width=0.16\textwidth]{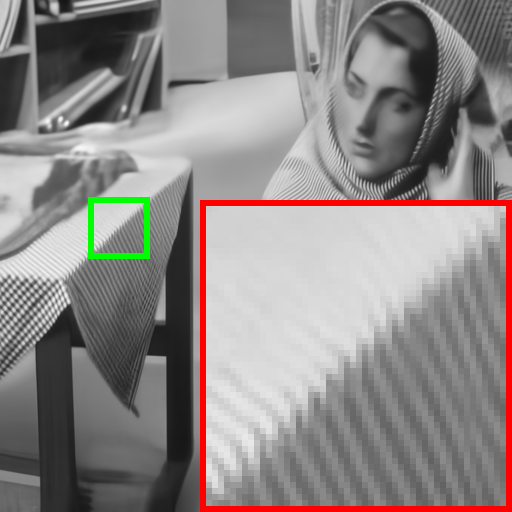}~
		&\includegraphics[width=0.16\textwidth]{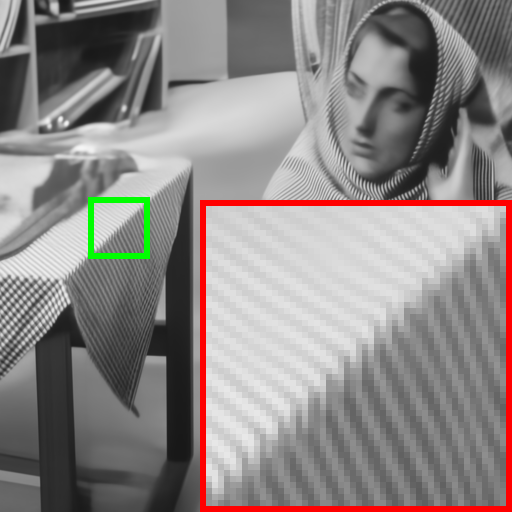}\\
 27.76/0.8222 &27.60/0.8243 & 28.29/0.8420  & 28.16/0.8423 & 28.41/0.8480 & 28.62/0.8560 \\
(g) NLRN & (h) FOCNet  & (i) DAGL & (j) DRUNet & (k) SwinIR &  (l) \textbf{SCUNet} \\
\end{tabular}
\vspace{-0.1cm}
\caption{Grayscale image denoising results of different methods on image ``\textit{Barbara}'' from Set12 dataset. The noisy image is corrupted by AWGN with noise level 50.}
\label{fig:grayscale}
\end{figure*}

\begin{spacing}{1.5}
\end{spacing}
\noindent
In practice, real images might be resized or JPEG compressed several times~\cite{jiang2021towards}, and JPEG compression might be performed before or after resizing. Inspired by this, our final degradation sequence employs a double degradation strategy and a random shuffle strategy. 
By doing this, the degradation space is expected to be largely expanded, which can facilitate the generalization ability of the trained deep blind model.
Specifically, we perform the above noises and resizing twice. We add
Gaussian noise and JPEG compression noise with the probabilities of 1.
For the resizing and other noise addition, we set the probabilities to 0.5.
Before applying the degradation sequence to a clean image, we first perform a random shuffle on the degradations. To prevent out-of-range values after each degradation process, we always make sure the image is clipped into the range of 0-1.
Due to the introduction of resizing, a large high quality image should be used for the paired training data synthesis. Fig.~\ref{degradation} provides a schematic illustration of the proposed training data synthesis pipeline.

\section{Discussion}

\subsection{Our denoising data synthesis pipeline \emph{vs.} super-resolution data synthesis pipeline~\cite{zhang2021designing,wang2021real}}

Our training data synthesis pipeline differs from the ones proposed in~\cite{zhang2021designing,wang2021real} in at least three main aspects. \textbf{First}, the applications are different. Our pipeline is used for deep blind image denoising, whereas the ones proposed in~\cite{zhang2021designing,wang2021real} are designed for deep blind super-resolution. 
\textbf{Second}, our pipeline also performs the resizing on the high-quality image to produce the corresponding clean image of the noisy images, whereas the degradation models in~\cite{zhang2021designing,wang2021real} do not perform such a procedure. The reason is that denoising does not necessitate removing image blur and enlarging the resolution, which is different from super-resolution.
\textbf{Third}, our pipeline adopts more kinds of noise, such as speckle noise. 
Fig.~\ref{fig:syn} shows some synthesized noisy/clean patch pairs via our proposed training data synthesis pipeline. It can be seen that our data synthesis pipeline can produce very realistic noisy images. It is worth noting that the noisy/clean patch pairs are from the same high quality image with size 544$\times$544. Since we also perform the resizing operations for clean image patches, we can observe some blurriness from some of the clean image patches.

\subsection{Practical blind denoising \emph{v.s} blind Gaussian denoising and blind camera sensor noise removal for DND and
SIDD}

Our practical blind denoising \emph{\textbf{is much more difficult}} than blind Gaussian denoising and blind camera sensor noise removal for DND and SIDD, and is the ``\emph{\textbf{true}}'' blind image denoising for practical application.
It is widely-known that the deep model trained for blind Gaussian denoising does not perform well for real images due to noise assumption mismatch. For this reason, DND and SIDD are established by capturing noisy and clean images pairs from different cameras. Although these two datasets help researchers shift to real image denoising, however, \textbf{they focus on camera sensor noise which also deviates significantly from the noise from the Internet in our daily life}. Moreover, as shown in Fig. 5, the state-of-the-art DeamNet for these datasets even has a worse result than Noise Clinic for noisy images from a different kind of camera, which indicates that \textbf{deep models trained for these two datasets do not generalize well for unseen noise, thus having very limited applications}. In contrast, our model is trained on a much more complex degradation model whose the degradation space is large enough to cover a large variety of different noise combinations, and thus can significantly improve the practicability.
As far as we know, the existing ``\emph{\textbf{true}}'' blind denoising is the work entitled ``\emph{The noise clinic: a blind image denoising algorithm}.'' \textbf{Our model can significantly outperform Noise Clinic and is the first deep model that can be readily applied for real applications}.

\begin{table*}[!htbp]\footnotesize
\center
\begin{center}
\caption{Average PSNR(dB) results of different methods for color image denoising with noise levels 15, 25 and 50 on the CBSD68, Kodak24, McMaster and Urban100 datasets. The best and second best results are highlighted in \textcolor[rgb]{1.00,0.00,0.00}{red} and \textcolor[rgb]{0.00,0.00,1.00}{blue} colors, respectively.}
\label{tab:denoising_color_results}
\begin{tabular}{p{0.8cm}<{\centering}p{0.8cm}<{\centering}|p{0.8cm}<{\centering}p{0.8cm}<{\centering}p{0.7cm}<{\centering}p{0.7cm}<{\centering}p{0.7cm}<{\centering}p{0.8cm}<{\centering}p{0.8cm}<{\centering}p{0.8cm}<{\centering}p{0.8cm}<{\centering}p{1cm}<{\centering}p{0.9cm}<{\centering}}
\toprule
\multirow{2}{*}{Dataset} & Noise &  \multirow{2}{*}{DnCNN}  &
\multirow{2}{*}{FFDNet} &
\multirow{2}{*}{DSNet} & 
\multirow{2}{*}{BRDNet} & 
\multirow{2}{*}{RNAN} &
\multirow{2}{*}{RDN} &
\multirow{2}{*}{IPT} &
\multirow{2}{*}{DRUNet} &
\multirow{2}{*}{SwinIR} &
\multirow{2}{*}{Restormer} &
\multirow{2}{*}{\textbf{SCUNet}}
\\
 & Level &  &   & &   & &  &  &  &    &    \\ \hline\hline
\multirow{3}{*}{CBSD68} & 15
& 33.90
& 33.87
& 33.91
& 34.10
& -
& -
& -
& {34.30}
& \textcolor[rgb]{1.00,0.00,0.00}{34.42}
& \textcolor[rgb]{0.00,0.00,1.00}{34.40}
& \textcolor[rgb]{0.00,0.00,1.00}{34.40}
\\
& 25
& 31.24 
& 31.21
& 31.28
& 31.43
& -
& -
& -
& 31.69
& \textcolor[rgb]{0.00,0.00,1.00}{31.78}
& \textcolor[rgb]{1.00,0.00,0.00}{31.79}
& \textcolor[rgb]{1.00,0.00,0.00}{31.79}
\\
& 50
& 27.95
& 27.96
& 28.05
& 28.16
& 28.27
& 28.31
& 28.39
& 28.51
& 28.56
& \textcolor[rgb]{0.00,0.00,1.00}{28.60}
& \textcolor[rgb]{1.00,0.00,0.00}{28.61}
\\
\hline
\multirow{3}{*}{Kodak24} & 15
& 34.60 
& 34.63
& 34.63
& 34.88
& -
& -
& -
& 35.31
& \textcolor[rgb]{0.00,0.00,1.00}{35.34}
& \textcolor[rgb]{1.00,0.00,0.00}{35.47}
& \textcolor[rgb]{0.00,0.00,1.00}{35.34}
\\
& 25
& 32.14
& 32.13
& 32.16
& 32.41
& -
& -
& -
& 32.89
& 32.89
& \textcolor[rgb]{1.00,0.00,0.00}{33.04}
& \textcolor[rgb]{0.00,0.00,1.00}{32.92}
\\
& 50
& 28.95
& 28.98
& 29.05
& 29.22
& 29.58
& 29.66
& 29.64
& 29.86
& {29.79}
& \textcolor[rgb]{1.00,0.00,0.00}{30.01}
& \textcolor[rgb]{0.00,0.00,1.00}{29.87}
\\
\hline
\multirow{3}{*}{McMaster} & 15
& 33.45
& 34.66
& 34.67
& 35.08
& -
& -
& -
& {35.40}
& \textcolor[rgb]{1.00,0.00,0.00}{35.61}
& \textcolor[rgb]{1.00,0.00,0.00}{35.61}
& \textcolor[rgb]{0.00,0.00,1.00}{35.60}
\\
& 25
& 31.52
& 32.35
& 32.40
& 32.75
& -
& -
& -
& {33.14}
& \textcolor[rgb]{0.00,0.00,1.00}{33.20}
& \textcolor[rgb]{1.00,0.00,0.00}{33.34}
& \textcolor[rgb]{1.00,0.00,0.00}{33.34}
\\
& 50
& 28.62 
& 29.18
& 29.28
& 29.52
& 29.72
& -
& 29.98
& {30.08}
& 30.22
& \textcolor[rgb]{1.00,0.00,0.00}{30.30}
& \textcolor[rgb]{0.00,0.00,1.00}{30.29}
\\
\hline
\multirow{3}{*}{Urban100} & 15
& 32.98
& 33.83
& -
& 34.42
& -
& -
& -
& {34.81}
& \textcolor[rgb]{0.00,0.00,1.00}{35.13}
& \textcolor[rgb]{0.00,0.00,1.00}{35.13}
& \textcolor[rgb]{1.00,0.00,0.00}{35.18}
\\
& 25
& 30.81
& 31.40
& -
& 31.99
& -
& -
& -
& {32.60}
& 32.90
& \textcolor[rgb]{0.00,0.00,1.00}{32.96}
& \textcolor[rgb]{1.00,0.00,0.00}{33.03}
\\
& 50
& 27.59
& 28.05
& -
& 28.56
& 29.08
& 29.38
& {29.71}
& {29.61}
& 29.82
& \textcolor[rgb]{0.00,0.00,1.00}{30.02}
& \textcolor[rgb]{1.00,0.00,0.00}{30.14}
\\
\bottomrule            
\end{tabular}
\end{center}
\end{table*}

\begin{figure*}[!htbp]\small
\hspace{-0.10cm}
\begin{tabular}{c@{\extracolsep{0em}}c@{\extracolsep{0.04em}}c@{\extracolsep{0.04em}}c@{\extracolsep{0.04em}}c@{\extracolsep{0.04em}}@{\extracolsep{0.04em}}c@{\extracolsep{0.04em}}c}
        \includegraphics[width=0.16\textwidth]{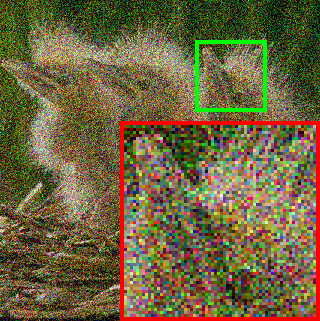}~
		&\includegraphics[width=0.16\textwidth]{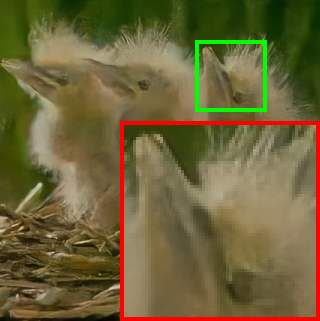}~
		&\includegraphics[width=0.16\textwidth]{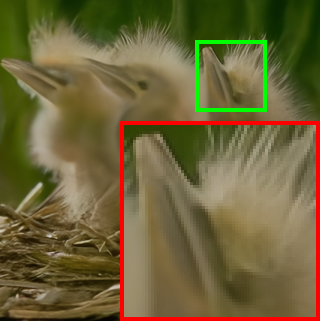}~
        &\includegraphics[width=0.16\textwidth]{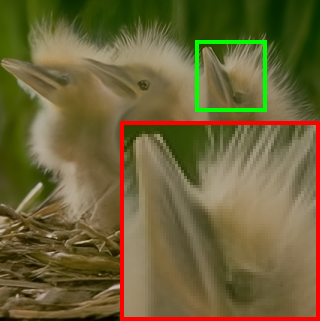}~
		&\includegraphics[width=0.16\textwidth]{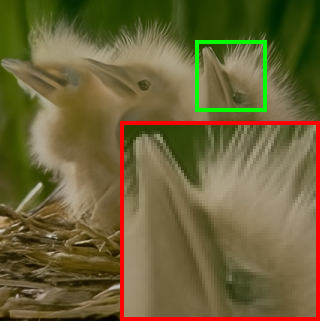}~
		&\includegraphics[width=0.16\textwidth]{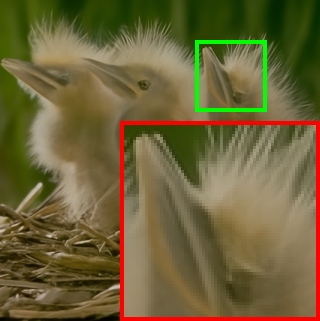}\\
 PSNR(dB)/SSIM &28.68/0.7794 & 29.06/0.7974  & \textcolor[rgb]{0.00,0.00,1.00}{29.28}/\textcolor[rgb]{0.00,0.00,1.00}{0.8103} & \textcolor[rgb]{0.00,0.00,1.00}{29.28}/0.8095 & \textcolor[rgb]{1.00,0.00,0.00}{29.37}/\textcolor[rgb]{1.00,0.00,0.00}{0.8135} \\
(a) Noisy & (b) DnCNN  & (c) RNAN  & (d) DRUNet & (e) SwinIR &  (f) \textbf{SCUNet} \\
	\end{tabular}
	\caption{Color image denoising results of different methods on image on image ``\textit{163085}'' from CBSD68 dataset. The noisy image is corrupted by AWGN with noise level 50.}
	\label{fig:color}
\end{figure*}

\begin{figure*}[!htbp]\small
\hspace{-0.17cm}
\begin{tabular}{c@{\extracolsep{0em}}c@{\extracolsep{0.04em}}c@{\extracolsep{0.04em}}c@{\extracolsep{0.04em}}c@{\extracolsep{0.04em}}@{\extracolsep{0.04em}}c@{\extracolsep{0.04em}}c}

        \includegraphics[width=0.16\textwidth]{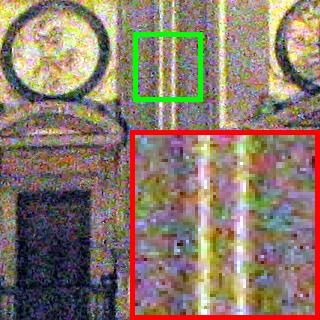}~
		&\includegraphics[width=0.16\textwidth]{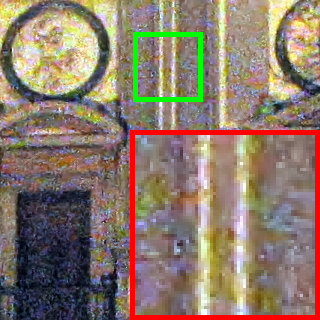}~
		&\includegraphics[width=0.16\textwidth]{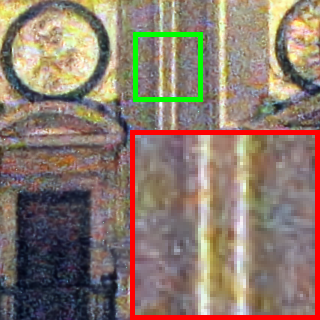}~
        &\includegraphics[width=0.16\textwidth]{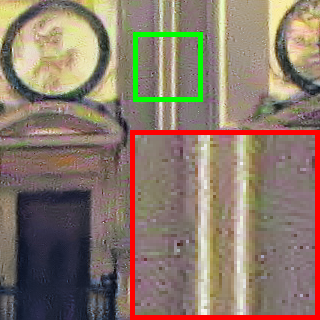}~
		&\includegraphics[width=0.16\textwidth]{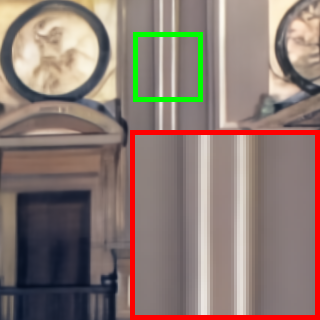}~
		&\includegraphics[width=0.16\textwidth]{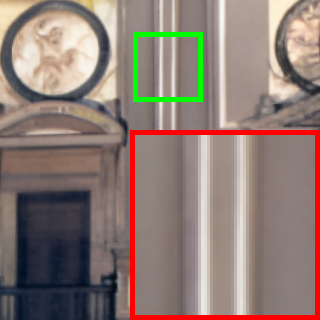}\\
5.53/8.40/10.11  & \textcolor[rgb]{0.00,0.00,1.00}{3.92}/\textcolor[rgb]{1.00,0.00,0.00}{8.94}/4.72  & \textcolor[rgb]{1.00,0.00,0.00}{3.78}/\textcolor[rgb]{0.00,0.00,1.00}{8.87}/\textcolor[rgb]{1.00,0.00,0.00}{3.99} & 5.53/8.40/\textcolor[rgb]{0.00,0.00,1.00}{4.16} & 6.64/3.89/58.75 & 6.63/3.69/55.85\\
        \includegraphics[width=0.16\textwidth]{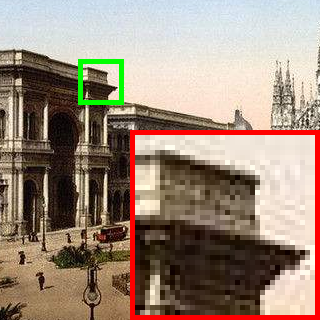}~
		&\includegraphics[width=0.16\textwidth]{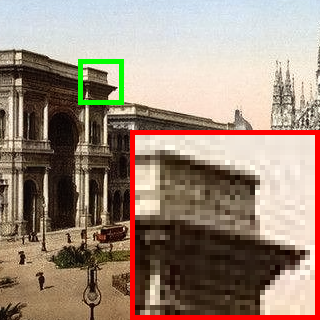}~
		&\includegraphics[width=0.16\textwidth]{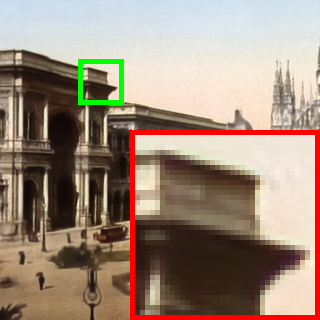}~
        &\includegraphics[width=0.16\textwidth]{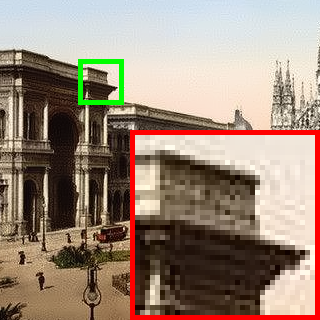}~
		&\includegraphics[width=0.16\textwidth]{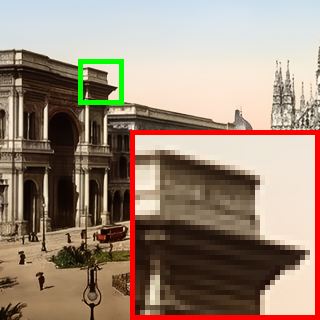}~
		&\includegraphics[width=0.16\textwidth]{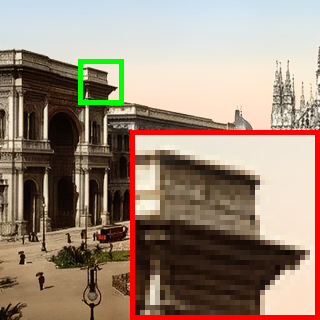}\\
\textcolor[rgb]{0.00,0.00,1.00}{4.60}/\textcolor[rgb]{0.00,0.00,1.00}{8.78}/11.69  & \textcolor[rgb]{1.00,0.00,0.00}{4.06}/\textcolor[rgb]{1.00,0.00,0.00}{8.80}/\textcolor[rgb]{0.00,0.00,1.00}{11.09}  & 4.92/6.91/20.49 & 4.61/\textcolor[rgb]{0.00,0.00,1.00}{8.78}/12.94 & 5.07/8.67/11.18 & 4.62/8.74/\textcolor[rgb]{1.00,0.00,0.00}{8.45} \\
        \includegraphics[width=0.16\textwidth]{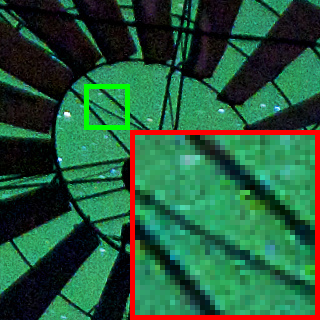}~
		&\includegraphics[width=0.16\textwidth]{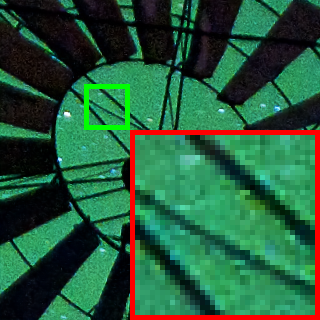}~
		&\includegraphics[width=0.16\textwidth]{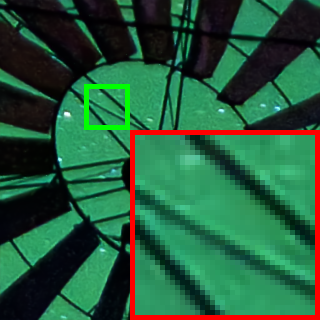}~
        &\includegraphics[width=0.16\textwidth]{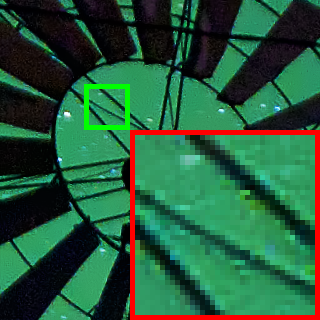}~
		&\includegraphics[width=0.16\textwidth]{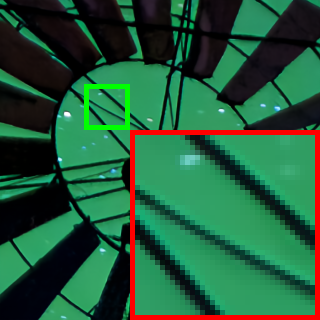}~
		&\includegraphics[width=0.16\textwidth]{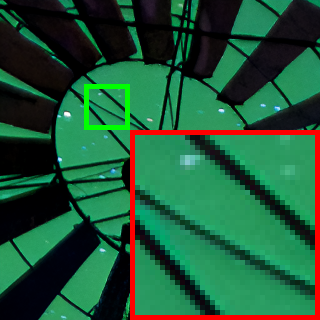}\\
\textcolor[rgb]{0.00,0.00,1.00}{4.50}/8.99/7.08  & \textcolor[rgb]{1.00,0.00,0.00}{3.93}/\textcolor[rgb]{1.00,0.00,0.00}{9.02}/\textcolor[rgb]{0.00,0.00,1.00}{6.99} & 5.51/7.85/28.78  & \textcolor[rgb]{0.00,0.00,1.00}{4.50}/\textcolor[rgb]{0.00,0.00,1.00}{9.00}/\textcolor[rgb]{1.00,0.00,0.00}{4.51} & 6.08/8.30/47.92 & 7.21/8.63/24.35 \\
(a) Noisy & (b) CBDNet  & (c) DeamNet  & (d) Noise Clinic & (e) \textbf{SCUNet}  &  (f) \textbf{SCUNetG} \\
	\end{tabular}
	\caption{Visual results and no-reference image quality assessment metrics (NIQE$\downarrow$/NRQM$\uparrow$/PIQE$\downarrow$) results of different methods for real image denoising. The images in each row from top to bottom are ``\textit{Palace}'', ``\textit{Building}'', and ``\textit{Stars}'', respectively.}
	\label{fig:real}
\end{figure*}
\section{Experiments}
\label{sec:experiments}

As discussed in Sec.~\ref{method}, the network architecture and the training data are two important factors to improve the performance of deep blind denoising model. For the sake of fairness, we first evaluate our SCUNet on synthetic Gaussian denoising. We then evaluate our training data synthesis pipeline with our SCUNet on practical blind image denoising.

\subsection{Synthetic Gaussian Denoising}
\label{sec:gaussiandenoising}

\paragraph{Implementation details.} For the high quality image dataset, we use the same training dataset consisting of Waterloo Exploration Database~\cite{ma2016gmad}, DIV2K~\cite{agustsson2017ntire}, and Flick2K~\cite{lim2017enhanced} for training. The settings of SwinT and Rconv blocks are same to those in SwinIR and DRUNet, respectively.
Following the common setting, we generate the noisy image by adding AWGN with a certain noise level and separately learn a denoising model for each noise level. The parameters are optimized by minimizing the L1 loss with Adam optimizer~\cite{kingma2014adam}.
The learning rate starts from 1e-4 and and decays by a factor of 0.5 every 200,000 iterations and finally ends with 3.125e-6. The patch size and batch size are set to 128$\times$128 and 24, respectively. We first train the model with noise level 25 and then finetune the model for other noise levels. All experiments are implemented by PyTorch 1.7.1. It takes about three days to train a denoising model on four NVIDIA RTX 2080 Ti GPUs.

\begin{spacing}{1.5}
\end{spacing}
\noindent
\textbf{Grayscale Gaussian denoising.} 
Table~\ref{table_denoising_grayscale} reports the PSNR results of different methods on the widely-used Set12~\cite{zhang2017beyond}, BSD68~\cite{MartinFTM01,roth2009fields}, Urban100~\cite{huang2015single} datasets for noise levels 15, 25 and 50. The compared methods include DnCNN~\cite{zhang2017beyond}, FFDNet~\cite{zhang2018ffdnet}, $\text{N}^3$Net~\cite{plotz2018neural}, NLRN~\cite{liu2018non}, RNAN~\cite{zhang2019rnan}, FOCNet~\cite{jia2019focnet}, DAGL~\cite{mou2021dynamic}, DRUNet~\cite{zhang2021plug}, SwinIR~\cite{liang2021swinir} and Restormer~\cite{zamir2022restormer}.
We note that $\text{N}^3$Net, NLRN, RNAN and SwinIR explicitly employ non-local module design in order to capture non-local image prior for better denoising performance. 
It can be seen that our SCUNet achieves significantly better PSNR results than other methods for all the noise levels on the three datasets. 
Specifically, SCUNet surpasses DnCNN and FFDNet by an average PSNR of 0.6dB on Set12, 0.3dB on BSD68 and 1.6dB on Urban100, and produces a substantial PSNR gain over state-of-the-art DAGL, DRUNet, SwinIR and Restormer. Since images from Urban100 are rich in repetitive structures, such a large improvement on Urban100 over BSD68 indicates that SCUNet is good at modeling non-local image prior.

To qualitatively evaluate the proposed SCUNet, we provide the denoising results of different methods on classical image ``\textit{Barbara}'' from Set12 dataset with noise level 50 in Fig.~\ref{fig:grayscale}. Note that we also include the traditional model-based methods BM3D~\cite{dabov2007image} and WNNM~\cite{gu2014weighted} for comparison since they are based on non-local priors. We have the following observations. First, WNNM produces much better visual results than some of the deep denoising methods such as DnCNN, FFDNet, RNAN and FOCNet. Second, while DAGL, DRUNet and SwinIR have better PSNR results than WNNM, they fail to recover some of the repetitive lines which indicates they still have limits in non-local prior modeling. Third, our SCUNet produces more visually pleasant results than others which further verifies the effectiveness of SCUNet for modeling image non-locality.

\begin{spacing}{1.5}
\end{spacing}
\noindent
\textbf{Color Gaussian denoising.} 
Table~\ref{tab:denoising_color_results} reports the color image denoising results of different methods on CBSD68~\cite{MartinFTM01,roth2009fields}, Kodak24~\cite{franzen1999kodak}, McMaster~\cite{zhang2011color} and Urban100~\cite{huang2015single} datasets. The compared methods include DnCNN, FFDNet, DSNet~\cite{peng2019dilated}, BRDNet~\cite{tian2020image}, RNAN, RDN~\cite{zhang2020residual}, IPT, DRUNet, SwinIR and Restormer.
As one can see, our SCUNet produces the best overall performance. Specifically, SCUNet surpasses DnCNN, FFDNet and DSNet by an average PSNR of 0.5dB on CBSD68, 0.7dB on Kodak24, 1.1dB on McMaster and 1.6dB on Urban100. Interestingly, while SCUNet has a similar PSNR gain over DRUNet for different noise levels, it achieves a larger PSNR gain than SwinIR with the increase of noise level. The possible reason is that SwinIR tends to lack the ability to model the long range dependency for heavy noise removal.

\begin{spacing}{1.5}
\end{spacing}
\noindent
\textbf{Results.}
Fig.~\ref{fig:real} provides the visual results of different blind denoising methods for real image denoising. The testing images includes ``\textit{Palace}'' from~\cite{lebrun2015noise}, ``\textit{Building}'' form Internet, and ``\textit{Stars}'' from~\cite{zhang2017beyond}.
The compared methods include CBDNet~\cite{guo2019toward}, DeamNet~\cite{ren2021adaptive} and Noise Clinic~\cite{lebrun2015noise}. We also report the results of no-reference image quality assessment (IQA) metrics NIQE~\cite{mittal2012making}, NRQM~\cite{ma2017learning} and PIQE~\cite{venkatanath2015blind}.

\begin{figure*}[!htbp]\footnotesize
\hspace{-0.12cm}
\begin{tabular}{c@{\extracolsep{0em}}@{\extracolsep{0.04em}}c@{\extracolsep{0.04em}}c@{\extracolsep{0.04em}}c@{\extracolsep{0.04em}}@{\extracolsep{0.04em}}c@{\extracolsep{0.04em}}c}
         \includegraphics[height=0.151\textwidth]{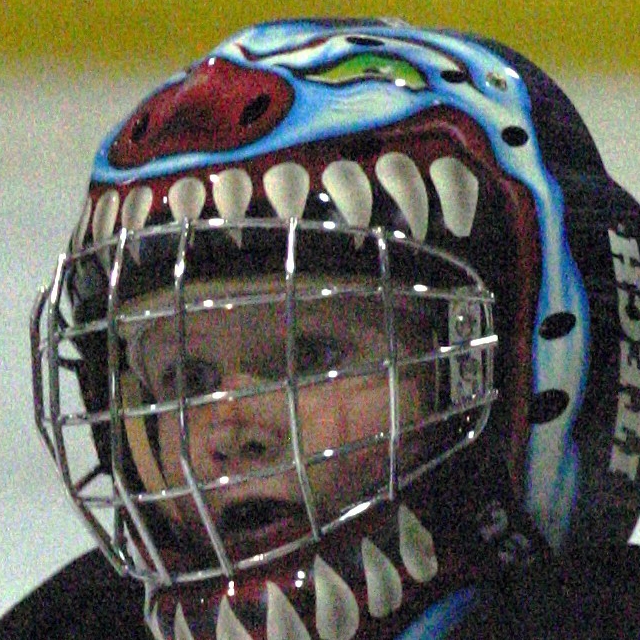}~
		&\includegraphics[height=0.151\textwidth]{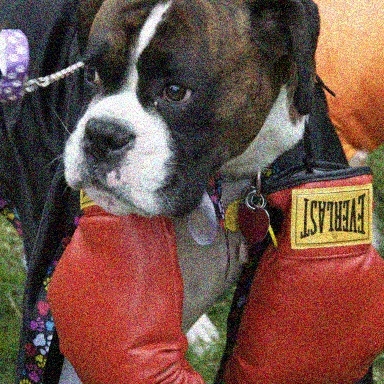}~
		&\includegraphics[height=0.151\textwidth]{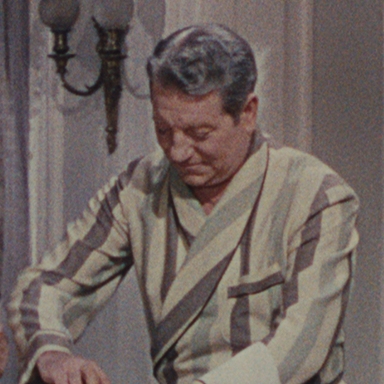}~
        &\includegraphics[height=0.151\textwidth]{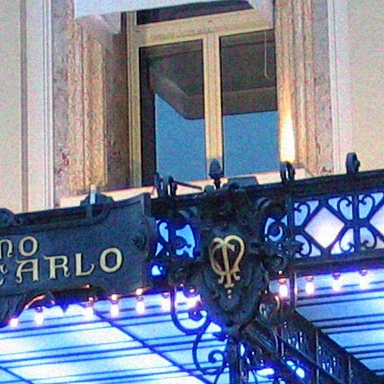}~
		&\includegraphics[height=0.151\textwidth]{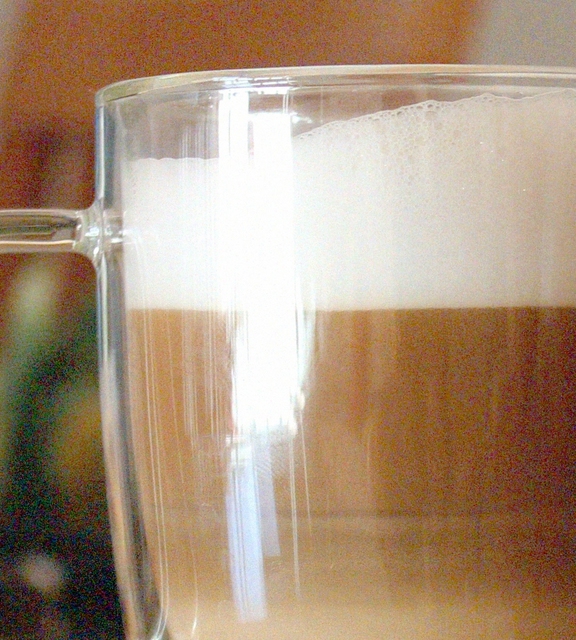}~
		&\includegraphics[height=0.151\textwidth]{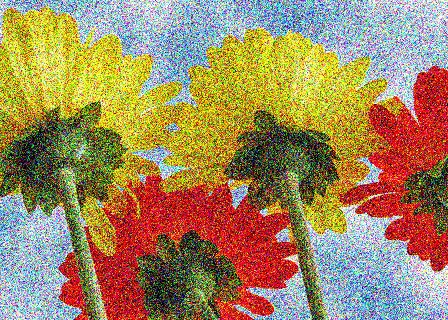}\\

         \includegraphics[height=0.151\textwidth]{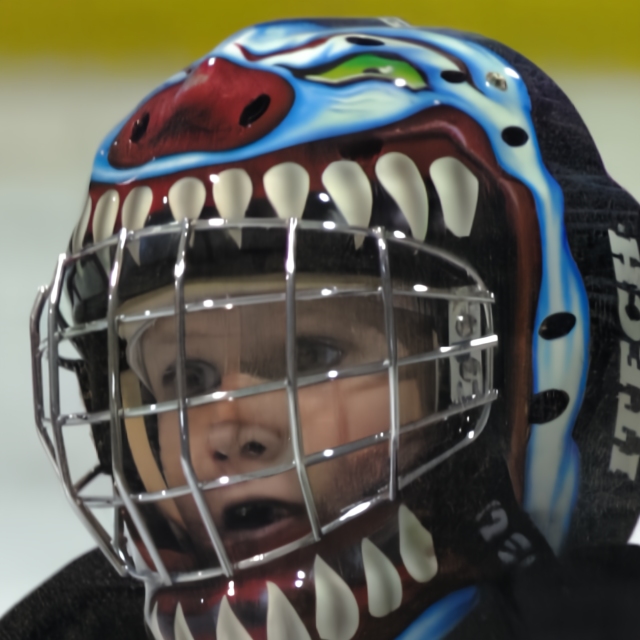}~
		&\includegraphics[height=0.151\textwidth]{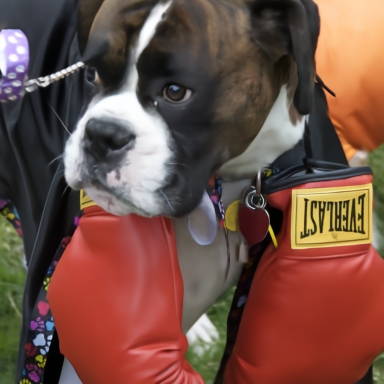}~
		&\includegraphics[height=0.151\textwidth]{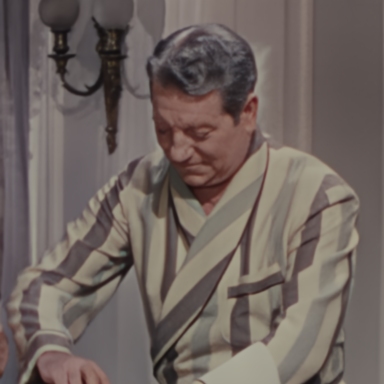}~
        &\includegraphics[height=0.151\textwidth]{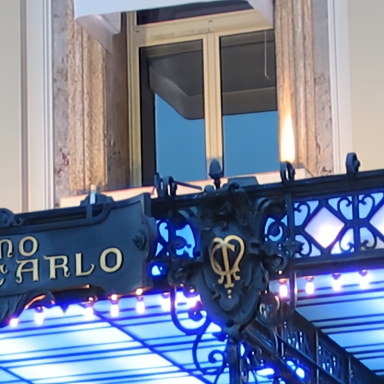}~
		&\includegraphics[height=0.151\textwidth]{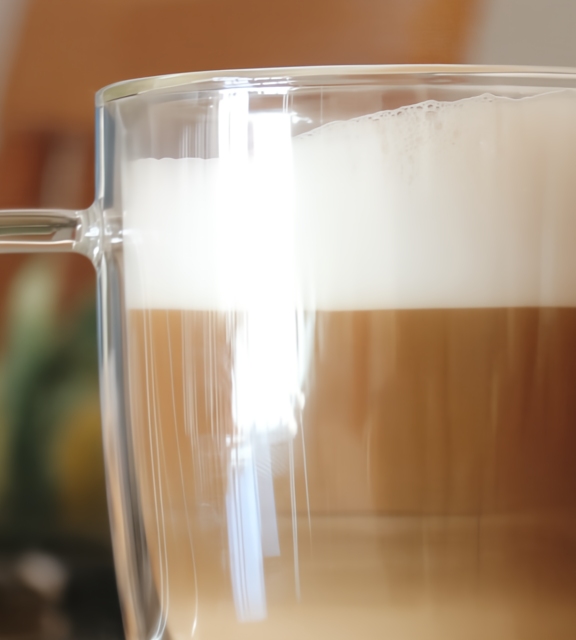}~
		&\includegraphics[height=0.151\textwidth]{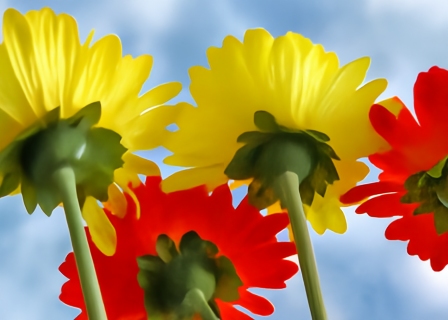}\\
		
         \includegraphics[height=0.151\textwidth]{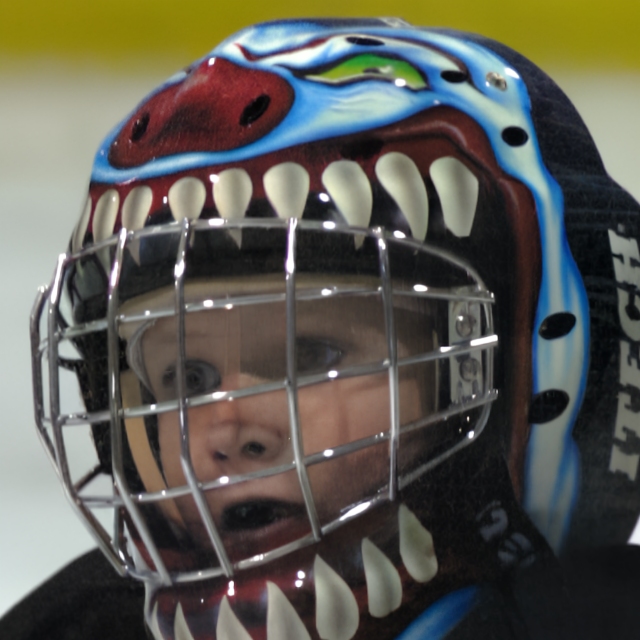}~
		&\includegraphics[height=0.151\textwidth]{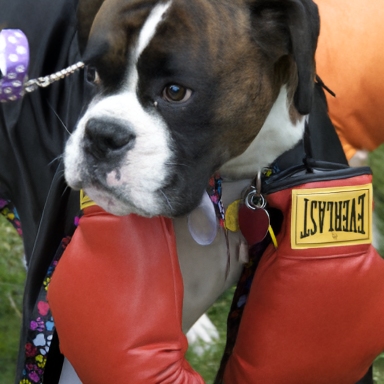}~
		&\includegraphics[height=0.151\textwidth]{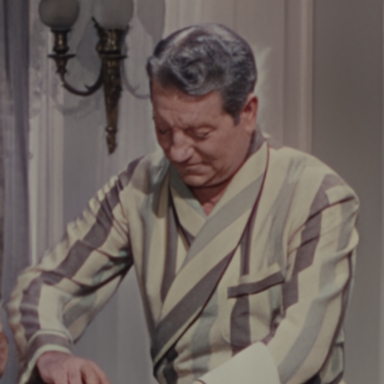}~
        &\includegraphics[height=0.151\textwidth]{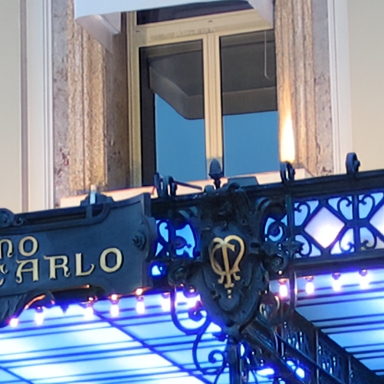}~
		&\includegraphics[height=0.151\textwidth]{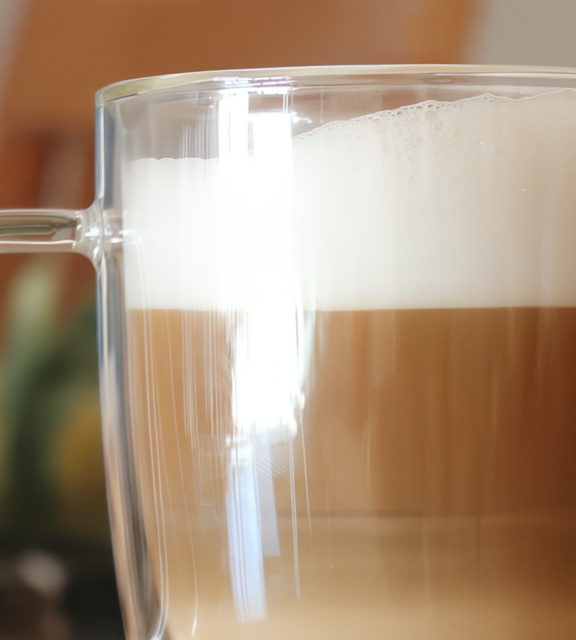}~
		&\includegraphics[height=0.151\textwidth]{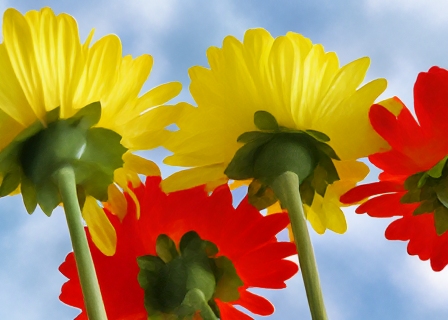}\\
(a) \emph{Boy} & (b) \emph{Dog}  & (c) \emph{Movie} & (d) \emph{Window} & (e) \emph{Glass} &  (f) \emph{Flowers} \\
	\end{tabular}
    \vspace{0.2cm}
	\caption{More blind denoising results of our SCUNet and SCUNetG on real images from RNI15 dataset. From top row to bottom row: noisy images, results of SCUNet, results of SCUNetG. Please zoom in for better view.}
	\label{fig:realsrset}
\end{figure*}

Fig.~\ref{fig:color} provides the visual results of different methods on image ``\textit{163085}'' from CBSD68 with noise level 50. It can be seen that SwinIR fails to recover the yellow structure along the beak of the bird while DnCNN, RNAN and DRUNet introduce some smoothness. By contrast, SCUNet recovers fine structures and preserves image sharpness.

\begin{table}[!htbp]\footnotesize
\vspace{-0.2cm}
\caption{FLOPs, runtime and \#Params comparisons on images of size 256$\times$256 on a PC with an Nvidia Titan Xp GPU.}
\center
\begin{tabular}{p{1.5cm}<{\centering}|p{1.4cm}<{\centering}p{1.4cm}<{\centering}p{1.5cm}<{\centering}}
 \toprule
Metrics & DRUNet& SwinIR & \textbf{SCUNet}  \\\hline\hline
FLOPs & \textcolor[rgb]{0.00,0.00,1.00}{143.5G}  & 787.9G & \textcolor[rgb]{1.00,0.00,0.00}{67.1G}  \\
Runtime &  \textcolor[rgb]{1.00,0.00,0.00}{0.020s}  & 0.525s & \textcolor[rgb]{0.00,0.00,1.00}{0.072s} \\
\#Params & 32.64M & \textcolor[rgb]{1.00,0.00,0.00}{11.49M} & \textcolor[rgb]{0.00,0.00,1.00}{17.94M} \\
\bottomrule
\end{tabular}
\label{table:runtime}
\end{table}

\begin{spacing}{1.5}
\end{spacing}
\noindent
\textbf{FLOPs, runtime, and \#Params.}
We report FLOPs, runtime and \#Params comparisons among DRUNet, SwinIR and SCUNet in Table~\ref{table:runtime}. We can see that our SCUNet achieves the lowest FLOPs due to the combination of UNet and SC block. Since SwinIR does not use any downscaling operations, it suffers from high FLOPs and long runtime. In comparison, SCUNet achieves the best trade-off between FLOPs, runtime and \#Params. Note that the runtime of SCUNet can be reduced by efficient implementation.

\subsection{Practical Blind Image Denoising}
\label{sec:realimagenoising}

\paragraph{Implementation details.} We use the same training implementations as in synthetic Gaussian denoising except the following: First, each high quality image is first cropped into a size of 544$\times$544 before processing it into a pair of noisy/clean images.
Second, the learning rate is fixed to 1e-4 as it tends to enhance the generalization ability. Third, we also train a perceptual quality-oriented blind model, namely SCUNetG, by minimizing a weighted combination of L1 loss, VGG perceptual loss on five convolution layers and UNetGAN loss~\cite{wang2021real} with weights $1$, $1$, and $1$, respectively.

\paragraph{Results.} From Fig.~\ref{fig:real}, we can observe that our SCUNet and SCUNetG achieve the best visual results for noise removal and details preserving. For example, both CBDNet and DeamNet fail to removal the processed camera sensor noise for ``\textit{Palace}'' while ours can remove such low-frequency noise and recover the underlying edges. However, our results do not show promising no-reference IQA results. As pointed out in~\cite{zhang2021designing}, such a phenomenon further indicates that no-reference IQA methods should update with degradation types. 
Fig.~\ref{fig:realsrset} provides more blind denoising results of our SCUNet and SCUNetG on real images from RNI15 dataset~\cite{zhang2018ffdnet}. 
Note that we do not know the ground-truth noise type and noise levels of these real images. For example, the ``\emph{Boy}'', ``\emph{Dog}'' and ``\emph{Glass}'' are likely to be corrupted by processed camera sensor noise with unknown camera type and the ``\emph{Flowers}'' is corrupted by Gaussian-like noise. Surprisingly, our models effectively handle these images, which could be due to the fact that they have been trained to manage a wide range of degradation scenarios created by various types of noise, resizing, and a random shuffle strategy.
According to the above results, we can conclude that the proposed training data synthesis pipeline is suitable for training deep blind denoising model for real applications. 

\begin{figure}[!htbp]\small
\begin{tabular}{c@{\extracolsep{0em}}c@{\extracolsep{0.04em}}c@{\extracolsep{0.04em}}c@{\extracolsep{0.04em}}c}
        \includegraphics[width=0.15\textwidth]{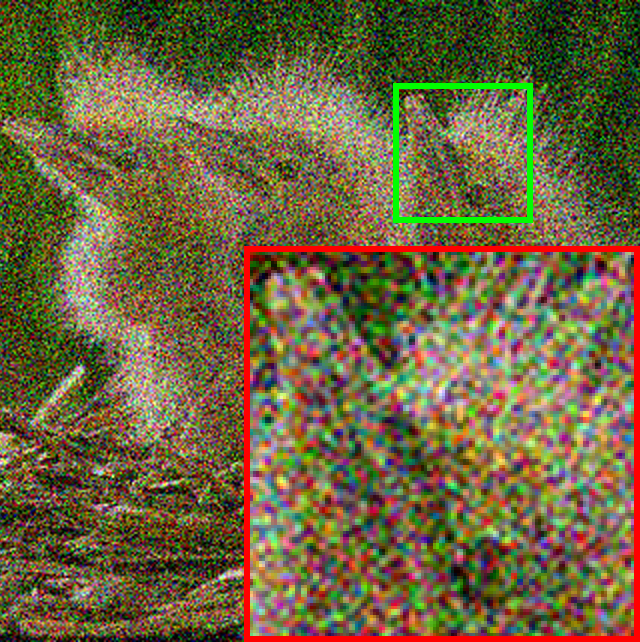}~
		&\includegraphics[width=0.15\textwidth]{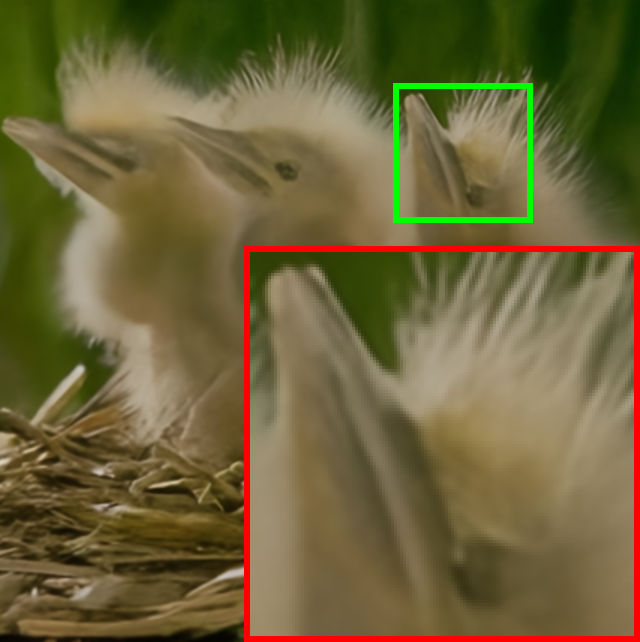}~
		&\includegraphics[width=0.15\textwidth]{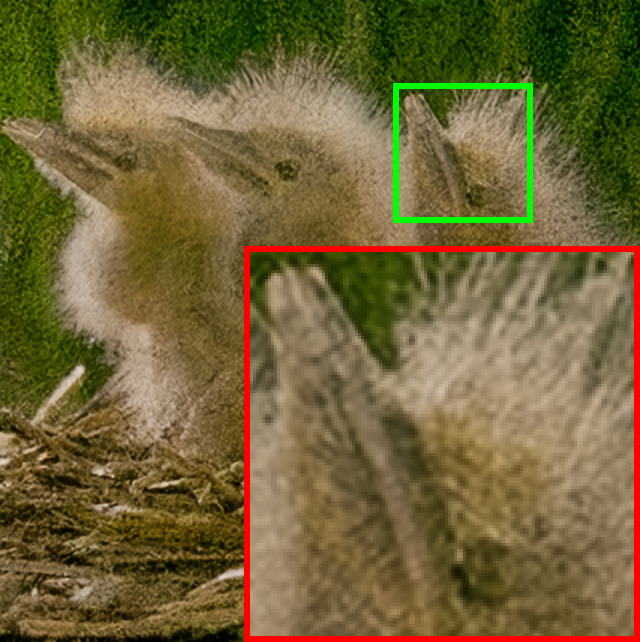}\\
        \includegraphics[width=0.15\textwidth]{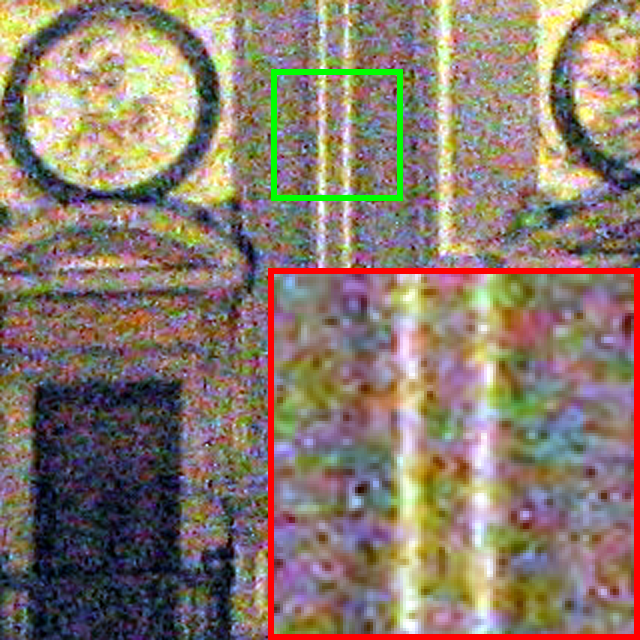}~
		&\includegraphics[width=0.15\textwidth]{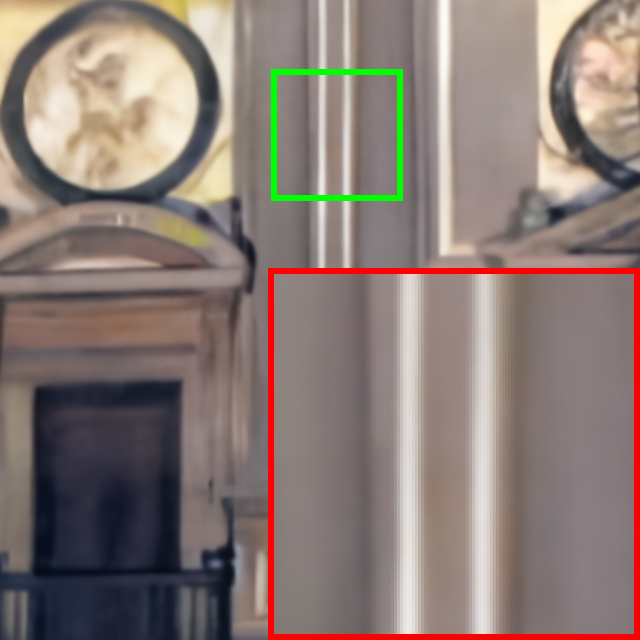}~
		&\includegraphics[width=0.15\textwidth]{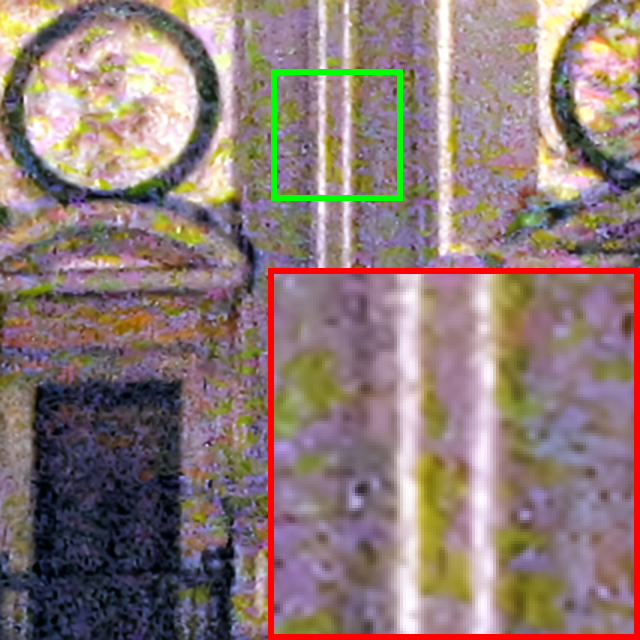}\\
(a) & (b)  &  (c) \\
	\end{tabular}
    \vspace{0.2cm}
	\caption{Comparison between SCUNet and its variant without using resizing in the training data synthesis for denoising a resized noisy image. (a) Upsampled noisy image by bicubic resizing with a scale factor of 2, (b) denoising result of SCUNet, (c) denoising results of SCUNet with using resizing in the training data synthesis.}
	\label{fig:resize}
\end{figure}

\paragraph{Impact of the resizing for data synthesis.}
Since one of the main differences between our proposed noisy image synthesis from others is that we adopt resizing to diversify the noise distribution, it is interesting to investigate the performance of the trained model without using resizing in the training data synthesis. Fig.~\ref{fig:resize} provides the visual comparisons on two upsampled noisy image by bicubic resizing with a scale factor of 2. The first noisy image is corrupted by Gaussian noise with noise level 50 while the second one is corrupted by unknown processed camera sensor noise. 
It can be seen that the trained model without using resizing in the training data synthesis fails to completely remove the noise. Thus, we can conclude that the resizing can help to improve the generalization ability.

\section{Conclusion}
\label{sec:conclusion}
In this paper, we focus on the problem of practical blind image denoising. Inspired by the Maximum A Posteriori (MAP) inference which indicates prior modeling and degradation modeling are essential for the success of deep blind denoising, we propose a new network architecture for better prior modeling and a novel data synthesis method for better practical usage. Specifically, we design a new  swin-conv block which incorporates the local modeling ability of residual convolution block and non-local modeling ability of swin transformer block, and plug it as the main building block into a UNet to further enhance the local and non-local modeling ability. Moreover, we design a data synthesis pipeline which considers different kinds of noise and also involves a random shuffle strategy and a double degradation strategy.
Extensive experimental results demonstrated the effectiveness of the new architecture design for Gaussian denoising and practicability of the trained deep blind model for real noisy images.

\section*{Acknowledgements}
This work was partly supported by the ETH Z\"urich Fund (OK), and by Huawei grant.


{\small
\bibliographystyle{ieee_fullname}
\bibliography{egbib}

\begin{thebibliography}{10}\itemsep=-1pt

\bibitem{afonso2010fast}
Manya~V Afonso, Jos{\'e}~M Bioucas-Dias, and M{\'a}rio~AT Figueiredo.
\newblock Fast image recovery using variable splitting and constrained
  optimization.
\newblock {\em IEEE transactions on image processing}, 19(9):2345--2356, 2010.

\bibitem{agustsson2017ntire}
Eirikur Agustsson and Radu Timofte.
\newblock Ntire 2017 challenge on single image super-resolution: Dataset and
  study.
\newblock In {\em IEEE Conference on Computer Vision and Pattern Recognition
  Workshops}, volume~3, pages 126--135, July 2017.

\bibitem{boyd2011distributed}
Stephen Boyd, Neal Parikh, Eric Chu, Borja Peleato, and Jonathan Eckstein.
\newblock Distributed optimization and statistical learning via the alternating
  direction method of multipliers.
\newblock {\em Foundations and Trends in Machine Learning}, 3(1):1--122, 2011.

\bibitem{brooks2019unprocessing}
Tim Brooks, Ben Mildenhall, Tianfan Xue, Jiawen Chen, Dillon Sharlet, and
  Jonathan~T Barron.
\newblock Unprocessing images for learned raw denoising.
\newblock In {\em IEEE Conference on Computer Vision and Pattern Recognition},
  pages 11036--11045, 2019.

\bibitem{buades2005non}
Antoni Buades, Bartomeu Coll, and Jean-Michel Morel.
\newblock A non-local algorithm for image denoising.
\newblock In {\em IEEE Conference on Computer Vision and Pattern Recognition},
  volume~2, pages 60--65, 2005.

\bibitem{burger2013learning}
Harold~Christopher Burger, Christian Schuler, and Stefan Harmeling.
\newblock Learning how to combine internal and external denoising methods.
\newblock In {\em German Conference on Pattern Recognition}, pages 121--130,
  2013.

\bibitem{cao2021swin}
Hu Cao, Yueyue Wang, Joy Chen, Dongsheng Jiang, Xiaopeng Zhang, Qi Tian, and
  Manning Wang.
\newblock Swin-unet: Unet-like pure transformer for medical image segmentation.
\newblock In {\em European Conference on Computer Vision Workshops}, pages
  205--218, 2023.

\bibitem{chatterjee2009denoising}
Priyam Chatterjee and Peyman Milanfar.
\newblock Is denoising dead?
\newblock {\em IEEE Transactions on Image Processing}, 19(4):895--911, 2009.

\bibitem{chen2021pre}
Hanting Chen, Yunhe Wang, Tianyu Guo, Chang Xu, Yiping Deng, Zhenhua Liu, Siwei
  Ma, Chunjing Xu, Chao Xu, and Wen Gao.
\newblock Pre-trained image processing transformer.
\newblock In {\em IEEE Conference on Computer Vision and Pattern Recognition},
  pages 12299--12310, 2021.

\bibitem{chen2018image}
Jingwen Chen, Jiawei Chen, Hongyang Chao, and Ming Yang.
\newblock Image blind denoising with generative adversarial network based noise
  modeling.
\newblock In {\em IEEE Conference on Computer Vision and Pattern Recognition},
  pages 3155--3164, 2018.

\bibitem{chen2015trainable}
Yunjin Chen and Thomas Pock.
\newblock Trainable nonlinear reaction diffusion: {A} flexible framework for
  fast and effective image restoration.
\newblock {\em IEEE transactions on Pattern Analysis and Machine Intelligence},
  39(6):1256--1272, 2016.

\bibitem{dabov2007image}
Kostadin Dabov, Alessandro Foi, Vladimir Katkovnik, and Karen Egiazarian.
\newblock Image denoising by sparse 3-{D} transform-domain collaborative
  filtering.
\newblock {\em IEEE Transactions on Image Processing}, 16(8):2080--2095, 2007.

\bibitem{franzen1999kodak}
Rich Franzen.
\newblock Kodak lossless true color image suite.
\newblock {\em source: http://r0k. us/graphics/kodak}, 4(2), 1999.

\bibitem{grossberg2003space}
Michael~D Grossberg and Shree~K Nayar.
\newblock What is the space of camera response functions?
\newblock In {\em IEEE Conference on Computer Vision and Pattern Recognition},
  pages II--602, 2003.

\bibitem{gu2014weighted}
Shuhang Gu, Lei Zhang, Wangmeng Zuo, and Xiangchu Feng.
\newblock Weighted nuclear norm minimization with application to image
  denoising.
\newblock In {\em IEEE Conference on Computer Vision and Pattern Recognition},
  pages 2862--2869, 2014.

\bibitem{guo2022cmt}
Jianyuan Guo, Kai Han, Han Wu, Yehui Tang, Xinghao Chen, Yunhe Wang, and Chang
  Xu.
\newblock {CMT}: Convolutional neural networks meet vision transformers.
\newblock In {\em IEEE Conference on Computer Vision and Pattern Recognition},
  pages 12175--12185, 2022.

\bibitem{guo2019toward}
Shi Guo, Zifei Yan, Kai Zhang, Wangmeng Zuo, and Lei Zhang.
\newblock Toward convolutional blind denoising of real photographs.
\newblock In {\em IEEE Conference on Computer Vision and Pattern Recognition},
  pages 1712--1722, 2019.

\bibitem{hasinoff2014photon}
Samuel~W Hasinoff.
\newblock Photon, poisson noise.

\bibitem{he2016identity}
Kaiming He, Xiangyu Zhang, Shaoqing Ren, and Jian Sun.
\newblock Identity mappings in deep residual networks.
\newblock {\em arXiv preprint arXiv:1603.05027}, 2016.

\bibitem{huang2015single}
Jia-Bin Huang, Abhishek Singh, and Narendra Ahuja.
\newblock Single image super-resolution from transformed self-exemplars.
\newblock In {\em IEEE Conference on Computer Vision and Pattern Recognition},
  pages 5197--5206, 2015.

\bibitem{jia2019focnet}
Xixi Jia, Sanyang Liu, Xiangchu Feng, and Lei Zhang.
\newblock Focnet: A fractional optimal control network for image denoising.
\newblock In {\em IEEE Conference on Computer Vision and Pattern Recognition},
  pages 6054--6063, 2019.

\bibitem{jiang2021towards}
Jiaxi Jiang, Kai Zhang, and Radu Timofte.
\newblock Towards flexible blind {JPEG} artifacts removal.
\newblock In {\em IEEE International Conference on Computer Vision}, pages
  4997--5006, 2021.

\bibitem{kamilov2017plug}
Ulugbek~S Kamilov, Hassan Mansour, and Brendt Wohlberg.
\newblock A plug-and-play priors approach for solving nonlinear imaging inverse
  problems.
\newblock {\em IEEE Signal Processing Letters}, 24(12):1872--1876, 2017.

\bibitem{kingma2014adam}
Diederik Kingma and Jimmy Ba.
\newblock Adam: A method for stochastic optimization.
\newblock In {\em International Conference for Learning Representations}, 2015.

\bibitem{krull2019noise2void}
Alexander Krull, Tim-Oliver Buchholz, and Florian Jug.
\newblock Noise2void-learning denoising from single noisy images.
\newblock In {\em IEEE Conference on Computer Vision and Pattern Recognition},
  pages 2129--2137, 2019.

\bibitem{lebrun2015noise}
Marc Lebrun, Miguel Colom, and Jean-Michel Morel.
\newblock The noise clinic: a blind image denoising algorithm.
\newblock {\em Image Processing On Line}, 5:1--54, 2015.

\bibitem{lefkimmiatis2017non}
Stamatios Lefkimmiatis.
\newblock Non-local color image denoising with convolutional neural networks.
\newblock In {\em IEEE Conference on Computer Vision and Pattern Recognition},
  pages 3587--3596, 2017.

\bibitem{li2021bossnas}
Changlin Li, Tao Tang, Guangrun Wang, Jiefeng Peng, Bing Wang, Xiaodan Liang,
  and Xiaojun Chang.
\newblock Bossnas: Exploring hybrid cnn-transformers with block-wisely
  self-supervised neural architecture search.
\newblock In {\em IEEE International Conference on Computer Vision}, pages
  12281--12291, 2021.

\bibitem{li2021localvit}
Yawei Li, Kai Zhang, Jiezhang Cao, Radu Timofte, and Luc Van~Gool.
\newblock {LocalViT}: Bringing locality to vision transformers.
\newblock {\em arXiv preprint arXiv:2104.05707}, 2021.

\bibitem{liang2021swinir}
Jingyun Liang, Jiezhang Cao, Guolei Sun, Kai Zhang, Luc Van~Gool, and Radu
  Timofte.
\newblock Swinir: Image restoration using swin transformer.
\newblock In {\em IEEE International Conference on Computer Vision Workshops},
  pages 1833--1844, 2021.

\bibitem{lim2017enhanced}
Bee Lim, Sanghyun Son, Heewon Kim, Seungjun Nah, and Kyoung Mu~Lee.
\newblock Enhanced deep residual networks for single image super-resolution.
\newblock In {\em IEEE Conference on Computer Vision and Pattern Recognition
  Workshops}, pages 136--144, 2017.

\bibitem{liu2018non}
Ding Liu, Bihan Wen, Yuchen Fan, Chen~Change Loy, and Thomas~S Huang.
\newblock Non-local recurrent network for image restoration.
\newblock In {\em Advances in Neural Information Processing Systems}, pages
  1673--1682, 2018.

\bibitem{liu2021swin}
Ze Liu, Yutong Lin, Yue Cao, Han Hu, Yixuan Wei, Zheng Zhang, Stephen Lin, and
  Baining Guo.
\newblock Swin transformer: Hierarchical vision transformer using shifted
  windows.
\newblock In {\em IEEE International Conference on Computer Vision}, 2021.

\bibitem{ma2017learning}
Chao Ma, Chih-Yuan Yang, Xiaokang Yang, and Ming-Hsuan Yang.
\newblock Learning a no-reference quality metric for single-image
  super-resolution.
\newblock {\em Computer Vision and Image Understanding}, 158:1--16, 2017.

\bibitem{ma2016gmad}
Kede Ma, Zhengfang Duanmu, Qingbo Wu, Zhou Wang, Hongwei Yong, Hongliang Li,
  and Lei Zhang.
\newblock Waterloo exploration database: New challenges for image quality
  assessment models.
\newblock {\em IEEE Transactions on Image Processing}, 26(2):1004--1016, 2017.

\bibitem{mairal2009non}
Julien Mairal, Francis Bach, Jean Ponce, Guillermo Sapiro, and Andrew
  Zisserman.
\newblock Non-local sparse models for image restoration.
\newblock In {\em IEEE International Conference on Computer Vision}, pages
  2272--2279, 2009.

\bibitem{MartinFTM01}
D. Martin, C. Fowlkes, D. Tal, and J. Malik.
\newblock A database of human segmented natural images and its application to
  evaluating segmentation algorithms and measuring ecological statistics.
\newblock In {\em IEEE International Conference on Computer Vision}, volume~2,
  pages 416--423, July 2001.

\bibitem{mittal2012making}
Anish Mittal, Rajiv Soundararajan, and Alan~C Bovik.
\newblock Making a ``completely blind'' image quality analyzer.
\newblock {\em IEEE Signal Processing Letters}, 20(3):209--212, 2012.

\bibitem{mou2021dynamic}
Chong Mou, Jian Zhang, and Zhuoyuan Wu.
\newblock Dynamic attentive graph learning for image restoration.
\newblock In {\em IEEE International Conference on Computer Vision}, pages
  4328--4337, 2021.

\bibitem{nam2016holistic}
Seonghyeon Nam, Youngbae Hwang, Yasuyuki Matsushita, and Seon~Joo Kim.
\newblock A holistic approach to cross-channel image noise modeling and its
  application to image denoising.
\newblock In {\em IEEE Conference on Computer Vision and Pattern Recognition},
  pages 1683--1691, 2016.

\bibitem{peng2019dilated}
Yali Peng, Lu Zhang, Shigang Liu, Xiaojun Wu, Yu Zhang, and Xili Wang.
\newblock Dilated residual networks with symmetric skip connection for image
  denoising.
\newblock {\em Neurocomputing}, 345:67--76, 2019.

\bibitem{plotz2017benchmarking}
Tobias Plotz and Stefan Roth.
\newblock Benchmarking denoising algorithms with real photographs.
\newblock In {\em Proceedings of the IEEE conference on computer vision and
  pattern recognition}, pages 1586--1595, 2017.

\bibitem{plotz2018neural}
Tobias Pl{\"o}tz and Stefan Roth.
\newblock Neural nearest neighbors networks.
\newblock In {\em Advances in Neural Information Processing Systems}, pages
  1087--1098, 2018.

\bibitem{racine1999speckle}
Ren{\'e} Racine, Gordon~AH Walker, Daniel Nadeau, Ren{\'e} Doyon, and Christian
  Marois.
\newblock Speckle noise and the detection of faint companions.
\newblock {\em Publications of the Astronomical Society of the Pacific},
  111(759):587, 1999.

\bibitem{ren2021adaptive}
Chao Ren, Xiaohai He, Chuncheng Wang, and Zhibo Zhao.
\newblock Adaptive consistency prior based deep network for image denoising.
\newblock In {\em IEEE Conference on Computer Vision and Pattern Recognition},
  pages 8596--8606, 2021.

\bibitem{ronneberger2015u}
Olaf Ronneberger, Philipp Fischer, and Thomas Brox.
\newblock U-net: Convolutional networks for biomedical image segmentation.
\newblock In {\em International Conference on Medical Image Computing and
  Computer-Assisted Intervention}, pages 234--241, 2015.

\bibitem{roth2009fields}
Stefan Roth and Michael~J Black.
\newblock Fields of experts.
\newblock {\em International Journal of Computer Vision}, 82(2):205--229, 2009.

\bibitem{schmidt2014shrinkage}
Uwe Schmidt and Stefan Roth.
\newblock Shrinkage fields for effective image restoration.
\newblock In {\em IEEE Conference on Computer Vision and Pattern Recognition},
  pages 2774--2781, 2014.

\bibitem{sun2011learning}
Jian Sun and Marshall~F Tappen.
\newblock Learning non-local range markov random field for image restoration.
\newblock In {\em IEEE Conference on Computer Vision and Pattern Recognition},
  pages 2745--2752, 2011.

\bibitem{tian2020image}
Chunwei Tian, Yong Xu, and Wangmeng Zuo.
\newblock Image denoising using deep cnn with batch renormalization.
\newblock {\em Neural Networks}, 121:461--473, 2020.

\bibitem{tur1982speckle}
Moshe Tur, Kuen-Chang Chin, and Joseph~W Goodman.
\newblock When is speckle noise multiplicative?
\newblock {\em Applied optics}, 21(7):1157--1159, 1982.

\bibitem{venkatanath2015blind}
N Venkatanath, D Praneeth, Maruthi~Chandrasekhar Bh, Sumohana~S Channappayya,
  and Swarup~S Medasani.
\newblock Blind image quality evaluation using perception based features.
\newblock In {\em Twenty First National Conference on Communications}, pages
  1--6, 2015.

\bibitem{wang2021real}
Xintao Wang, Liangbin Xie, Chao Dong, and Ying Shan.
\newblock {Real-ESRGAN}: Training real-world blind super-resolution with pure
  synthetic data.
\newblock In {\em IEEE International Conference on Computer Vision Workshops},
  2021.

\bibitem{wang2021uformer}
Zhendong Wang, Xiaodong Cun, Jianmin Bao, Wengang Zhou, Jianzhuang Liu, and
  Houqiang Li.
\newblock Uformer: A general u-shaped transformer for image restoration.
\newblock In {\em IEEE Conference on Computer Vision and Pattern Recognition},
  pages 17683--17693, 2022.

\bibitem{yuan2021incorporating}
Kun Yuan, Shaopeng Guo, Ziwei Liu, Aojun Zhou, Fengwei Yu, and Wei Wu.
\newblock Incorporating convolution designs into visual transformers.
\newblock In {\em IEEE International Conference on Computer Vision}, pages
  579--588, 2021.

\bibitem{yue2019variational}
Zongsheng Yue, Hongwei Yong, Qian Zhao, Lei Zhang, and Deyu Meng.
\newblock Variational denoising network: Toward blind noise modeling and
  removal.
\newblock In {\em NeurIPS}, 2019.

\bibitem{zamir2022restormer}
Syed~Waqas Zamir, Aditya Arora, Salman Khan, Munawar Hayat, Fahad~Shahbaz Khan,
  and Ming-Hsuan Yang.
\newblock Restormer: Efficient transformer for high-resolution image
  restoration.
\newblock In {\em IEEE Conference on Computer Vision and Pattern Recognition},
  pages 5728--5739, 2022.

\bibitem{zhang2021plug}
Kai Zhang, Yawei Li, Wangmeng Zuo, Lei Zhang, Luc Van~Gool, and Radu Timofte.
\newblock Plug-and-play image restoration with deep denoiser prior.
\newblock {\em IEEE Transactions on Pattern Analysis and Machine Intelligence},
  2021.

\bibitem{zhang2021designing}
Kai Zhang, Jingyun Liang, Luc Van~Gool, and Radu Timofte.
\newblock Designing a practical degradation model for deep blind image
  super-resolution.
\newblock In {\em IEEE International Conference on Computer Vision}, pages
  4791--4800, 2021.

\bibitem{zhang2017beyond}
Kai Zhang, Wangmeng Zuo, Yunjin Chen, Deyu Meng, and Lei Zhang.
\newblock Beyond a gaussian denoiser: Residual learning of deep {CNN} for image
  denoising.
\newblock {\em IEEE Transactions on Image Processing}, pages 3142--3155, 2017.

\bibitem{zhang2018ffdnet}
Kai Zhang, Wangmeng Zuo, and Lei Zhang.
\newblock {FFDNet}: Toward a fast and flexible solution for cnn-based image
  denoising.
\newblock {\em IEEE Transactions on Image Processing}, 27(9):4608--4622, 2018.

\bibitem{zhang2011color}
Lei Zhang, Xiaolin Wu, Antoni Buades, and Xin Li.
\newblock Color demosaicking by local directional interpolation and nonlocal
  adaptive thresholding.
\newblock {\em Journal of Electronic Imaging}, 20(2):023016, 2011.

\bibitem{zhang2019rnan}
Yulun Zhang, Kunpeng Li, Kai Li, Bineng Zhong, and Yun Fu.
\newblock Residual non-local attention networks for image restoration.
\newblock In {\em International Conference on Learning Representations}, 2019.

\bibitem{zhang2020residual}
Yulun Zhang, Yapeng Tian, Yu Kong, Bineng Zhong, and Yun Fu.
\newblock Residual dense network for image restoration.
\newblock {\em IEEE Transactions on Pattern Analysis and Machine Intelligence},
  43(7):2480--2495, 2020.

\end{thebibliography}
}

\end{document}